\documentclass[12pt]{article}
\usepackage{fullpage}
\usepackage{subfigure} 
\usepackage{amsmath,amsfonts,amssymb}
\usepackage[utf8]{inputenc} 
\usepackage[T1]{fontenc}    
\usepackage{hyperref}       
\usepackage{url}            
\usepackage{booktabs}       
\usepackage{amsfonts}       
\usepackage{nicefrac}       
\usepackage{microtype}      
\usepackage{graphicx}
\usepackage{tikz,pgfplots}
\usepackage{framed}
\usepackage{array}
\usepackage{placeins}

\usepackage{algorithm}
\usepackage{algorithmic}
\usepackage{amsmath}
\usepackage{amssymb}
\usepackage{amsthm}
\usepackage{color}
\usepackage{multirow}

\usepackage{multirow}
\usepackage{cellspace}
\usepackage{setspace}
\usepackage{bm}
\usepackage{bbm}
\usepackage{graphicx}
\usepackage{tikz,pgfplots}
\usepackage{framed}
\usepackage{tabularx,colortbl}
\usepackage{cellspace}

\usepackage[round]{natbib}

\usepackage{hyperref}
\hypersetup{colorlinks,
            linkcolor=blue,
            citecolor=blue,
            urlcolor=magenta,
            linktocpage,
            plainpages=false}

\usepackage[capitalise]{cleveref}

\newcommand{\startfoo}{%
    \par\medskip
    \begin{mdframed}[linewidth=1pt]%
    \let\figure\figurehere
    \let\endfigure\endfigurehere
    \ignorespaces
}
\newcommand{\stopfoo}{%
    \unskip
    \end{mdframed}%
    \par\medskip
}

\usepackage{anyfontsize}
\newcommand{\smaller}{\fontsize{9}{9}\selectfont}

\title{An Online Learning Approach to Generative Adversarial Networks}

\author{%
Paulina Grnarova\\
ETH Z\"urich\\
 \texttt{\smaller paulina.grnarova@inf.ethz.ch}
\and
Kfir Y. Levy \\
ETH Z\"urich\\ 
  \texttt{\smaller  yehuda.levy@inf.ethz.ch }
\and 
Aurelien Lucchi \\
ETH Z\"urich\\ 
  \texttt{\smaller aurelien.lucchi@inf.ethz.ch }
   \and 
 Thomas Hofmann \\
ETH Z\"urich\\ 
  \texttt{\smaller thomas.hofmann@inf.ethz.ch }
\and 
Andreas Krause \\
ETH Z\"urich\\ 
\texttt{\smaller krausea@ethz.ch }
}

\newcommand{\D}{{\mathcal D}}
\newcommand{\E}{\mathbb{E}}

\newcommand{\Q}{{\mathcal Q}}
\newcommand{\R}{\mathcal{R}}

\newcommand{\tg}{\tilde{g}}

\renewcommand{\u}{{\mathbf u}}
\renewcommand{\v}{{\mathbf v}}

\newcommand{\z}{{\mathbf z}}
\newcommand{\x}{{\mathbf x}}

\newcommand{\pdata}{p_{\text{data}}}

\newcommand{\FTRLGAN}{{\sc Chekhov GAN~}}

\newcommand{\argmax}{\mathop\mathrm{{argmax}}}
\newcommand{\argmin}{\mathop\mathrm{{argmin}}}

\newtheorem{theorem}{Theorem}
\newtheorem{lemma}{Lemma}

\newtheorem{proposition}{Proposition}
\newtheorem{definition}{Definition}

\newcommand{\A}{\mathcal{A}}
\renewcommand{\P}{\mathcal{P}}

\newcommand{\F}{\mathcal{F}}

\newcommand{\blr}[1]{\big(#1\big)}

\renewcommand{\O}{O}

\def\reals{{\mathcal R}}

\newcommand{\K}{\mathcal{K}}

\newcommand{\ignore}[1]{}

\def\reals{{\mathbb R}}

\def\bold0{\mathbf{0}}

\def\x{\mathbf{x}}

\def\z{\mathbf{z}}

\def\v{\mathbf{v}}

\newcommand{\eps}{\varepsilon}

\graphicspath{{./Images/}}
\newcolumntype{S}{>{\centering\arraybackslash} m{.10\linewidth} }
\newcolumntype{T}{>{\centering\arraybackslash} m{.30\linewidth} }


\begin{document}

\maketitle

\begin{abstract}
We consider the problem of training generative models with a Generative Adversarial Network (GAN). Although GANs can accurately model complex distributions, they are known to be difficult to train due to instabilities caused by a difficult minimax optimization problem. In this paper, we view the problem of training GANs as finding a \emph{mixed} strategy in a zero-sum game.
Building on ideas from online learning we propose a novel training method named \FTRLGAN\footnote{We base this name on the Chekhov's gun (dramatic) principle that states that every element in a story must be necessary, and irrelevant elements should be removed. Analogously, our \FTRLGAN algorithm  introduces a sequence of elements which are eventually composed to yield a generator.}.
On the theory side, we show that our method provably converges to an equilibrium for semi-shallow GAN architectures, i.e. architectures where the discriminator is a one layer network and the generator is arbitrary.
On the practical side, we develop an efficient heuristic guided by our theoretical results, which we apply to commonly used deep GAN architectures.
On several real world tasks our approach exhibits improved stability and performance compared to standard GAN training.
\end{abstract}
\newpage

\section{Introduction}

A recent trend in generative models is to use a deep neural network as a generator. Two notable approaches  are variational auto-encoders (VAE)~\cite{Kingma:2013tz, Rezende:2014vm} as well as  Generative Adversarial Networks (GAN)~\cite{Goodfellow:2014td}. 
Unlike VAEs, the GAN approach offers a way to circumvent log-likelihood-based estimation and it also typically produces visually sharper samples~\cite{Goodfellow:2014td}. The goal of the generator network is to generate samples that are indistinguishable from real samples, where indistinguishability is measured by an additional discriminative model. This creates an adversarial game setting where one pits a generator against a discriminator.

Let us denote the data distribution by $\pdata(\x)$ and the model distribution by $p_\u(\x)$. A probabilistic discriminator is denoted by $h_\v: \x \to [0;1]$ and a generator by $G_\u: \z \to \x$. The GAN objective is:
\begin{align}
\min_{\u} \max_{\v}M(\u,\v) &= \frac 12 \E_{\x \sim \pdata} \log h_\v(\x) + \frac 12 \E_{\z \sim p_\z} \log (1-h_\v(G_\u(\z)))~.
\label{eq:GAN_objective}
\end{align}

Each of the two players (generator/discriminator) tries to optimize their own objective, which is exactly balanced by the loss of the other player, thus yielding a two-player zero-sum minimax game. 
Standard GAN approaches aim at finding a pure Nash Equilibrium by using traditional gradient-based techniques to minimize each player’s cost in an alternating fashion. However, an update made by one player can repeatedly undo the progress made by the other one, without ever converging. 

In general, alternating gradient descent fails to converge even for very simple games ~\cite{salimans2016improved}. In the setting of GANs, one of the central open issues is this non-convergence problem, which in practice leads to oscillations between different kinds of generated samples~\cite{metz2016unrolled}. 

While standard GAN methods seek to find  pure minimax strategies, we propose to consider mixed strategies, which allows us to leverage online learning algorithms for mixed strategies in large games. Building on the approach of~\cite{freund1999adaptive}, we propose a novel training algorithm for GANs that we call \FTRLGAN. 

\begin{figure*}[t]
\centering
\subfigure[]{ \label{fig:shallow}
\includegraphics[trim = 20mm 65mm 85mm 47mm, clip,
width=0.31\textwidth ]{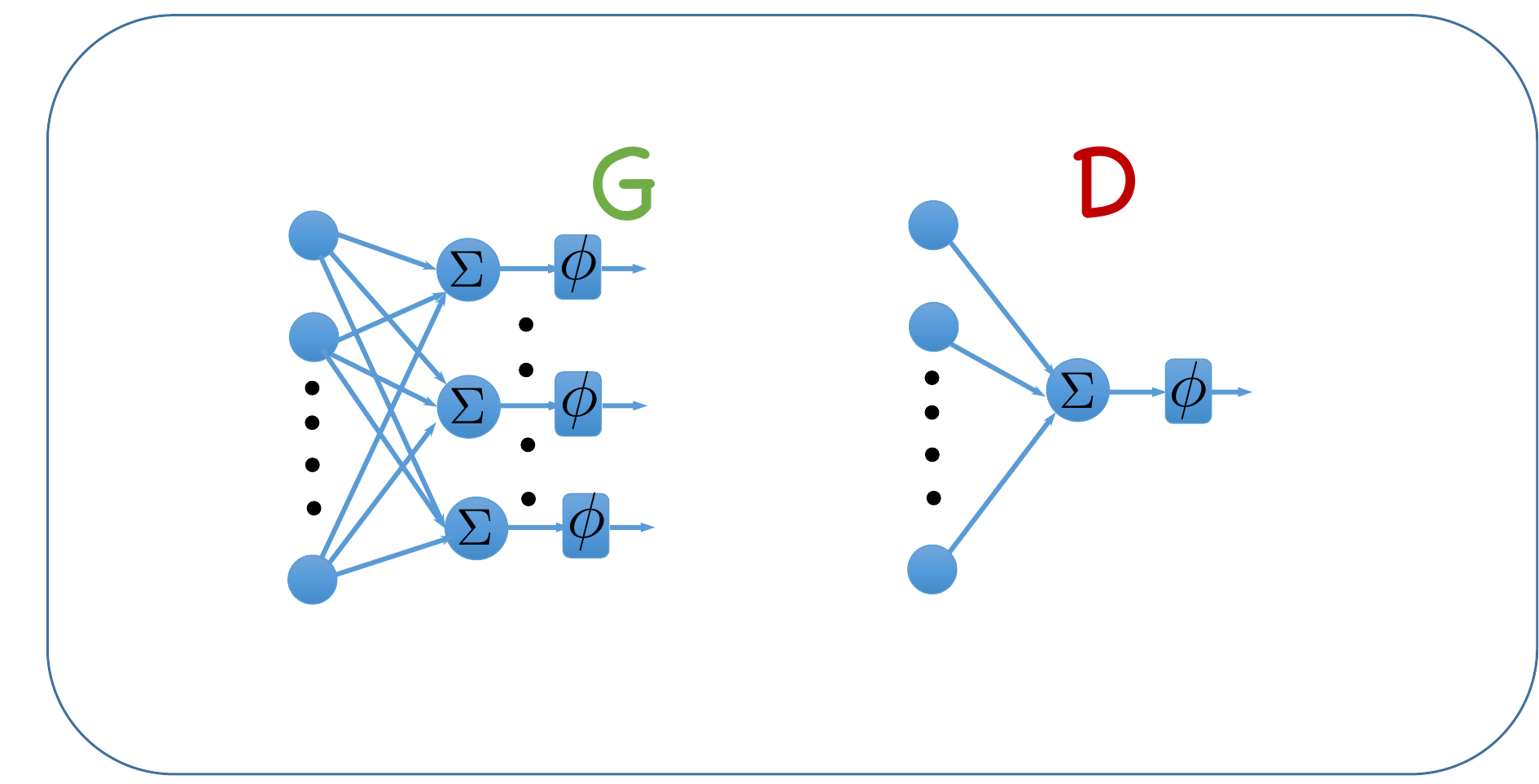}}
\subfigure[]{\label{fig:semishallow}
 \includegraphics[trim = 42mm 71mm 107mm 60mm, clip,
width=0.31\textwidth ]{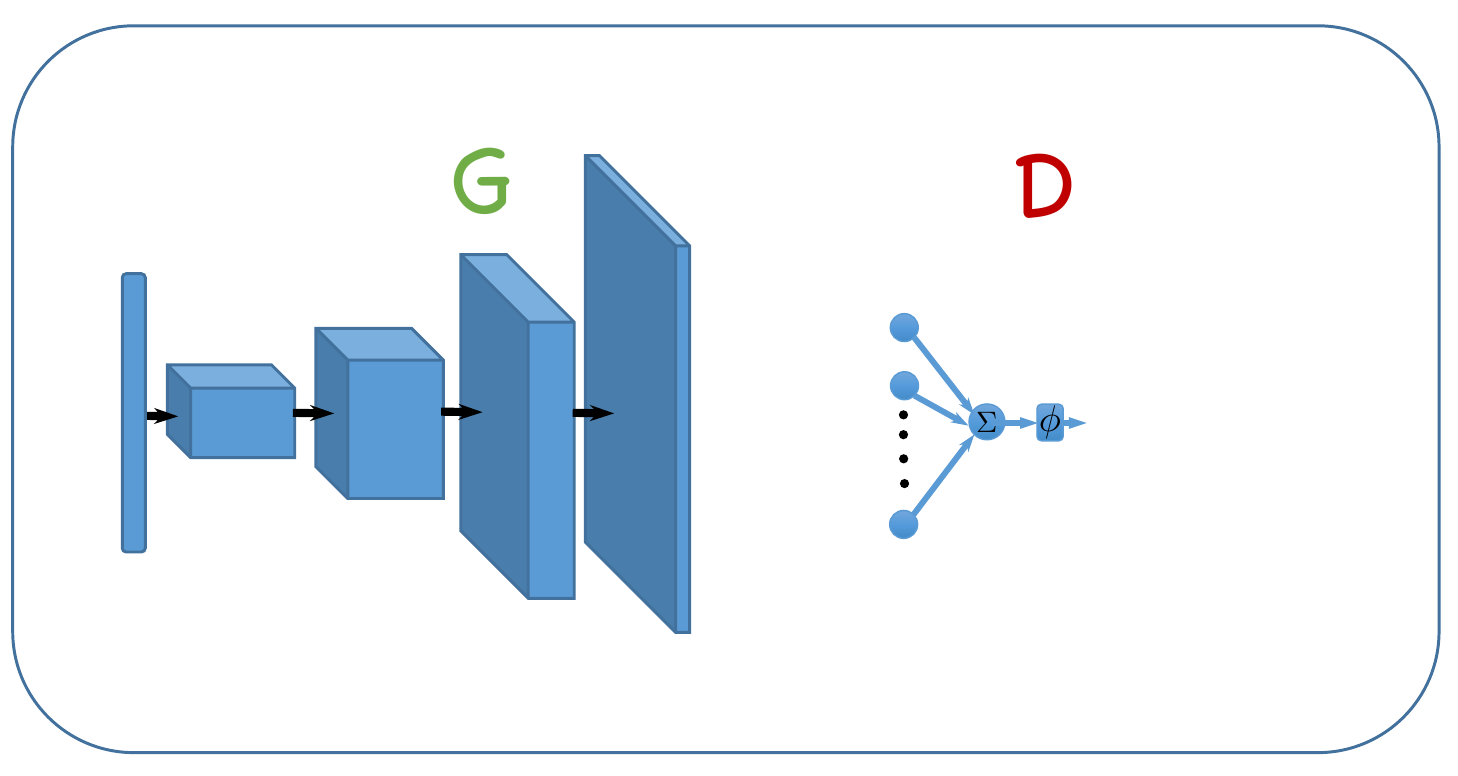}}
\subfigure[]{ \label{fig:deep}
 \includegraphics[trim =32mm 51mm 117mm 80mm, clip,
width=0.31\textwidth ]{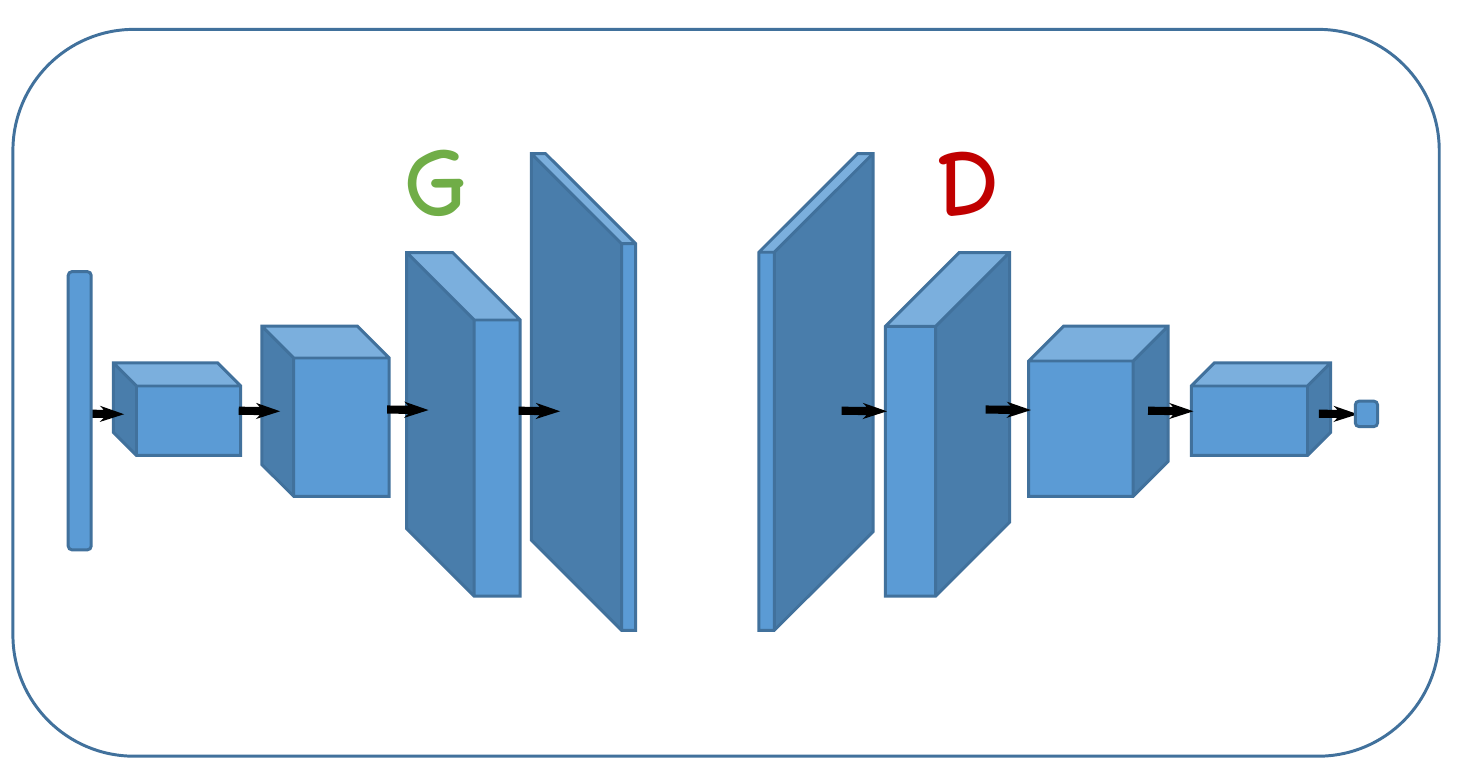}}
\caption{Three types of GAN architectures. Left: \textit{shallow}. Middle: \textit{semi-shallow}. Right: \textit{deep}.} 
\label{fig:GAN_architectures}
\end{figure*}

On the theory side, we focus on simpler GAN architectures. The most elementary architecture that one might consider is a \textit{shallow} one, e.g. a GAN architecture which consists of a single layer network as a discriminator, and a generator with one hidden layer (see Fig.~\ref{fig:GAN_architectures}). However, one typically requires a powerful generator that can model complex data distribution. This leads us to consider a \textit{semi-shallow} architecture where the generator is any arbitrary network (Fig.~\ref{fig:semishallow}). In this paper, we address the following questions: \textbf{1) Can we efficiently find an equilibrium for semi-shallow GAN architectures? 2) Can we extend this result to more complex architectures?}

We answer the first question in the affirmative, and provide a method that provably finds an equilibrium in the setting of a semi-shallow architecture. This is done in spite of the fact that the game induced by such architectures is \emph{not} convex-concave. Our proof relies on analyzing semi-concave games, i.e., games which are concave with respect to the $\max$ player, but need not have a special structure with respect to the $\min$ player.
We prove that in such games, players may efficiently invoke regret minimization procedures in order to find equilibrium. To the best of our knowledge, this result is novel in the context of GANs, and might also find use in other scenarios where such structure may arise.

On the practical side, we develop an efficient heuristic guided by our theoretical results, which we apply to commonly used \textit{deep} GAN architectures shown in Fig.~\ref{fig:deep}. We provide experimental results demonstrating that our approach exhibits better empirical stability compared to GANs and generates more diverse samples, while retaining the visual quality.

In Section~\ref{sec:background}, we briefly review necessary notions from online learning and zero-sum games. We then present our approach and its theoretical guarantees in Section~\ref{sec:GANs}. Lastly, we present empirical results on standard benchmark datasets in Section~\ref{sec:experiments}.

\section{Background \& Related Work}
\label{sec:background}

\subsection{GANs}
\textbf{GAN Objectives:}
The classical way to learn a generative model consists of minimizing a divergence function between a parametrized model distribution $p_\u(\x)$ and the true data distribution $\pdata(\x)$. The original GAN approach~\cite{Goodfellow:2014td} was shown to be related to the Jensen-Shannon divergence.
This was later generalized by~\cite{nowozin2016f} that described a broader family of GAN objectives stemming from  $f$-divergences. A different popular type of GAN objectives is the family of Integral Probability Metrics~\cite{muller1997integral}, such as the kernel MMD~\cite{gretton2012kernel, li2015generative} or the Wasserstein metric~\cite{arjovsky2017towards}. All of these divergence measures yield a minimax objective.\\

\textbf{Training methods for GANs:}
In order to solve the minimax objective in Eq.~\ref{eq:GAN_objective}, the authors of~\cite{Goodfellow:2014td} suggest an approach that alternatively minimizes over $\u$ and $\v$ using mini-batch stochastic gradient descent and show convergence when the updates are made in function space. In practice, this condition is not met - since this procedure works in the parameter space - and many issues arise during training~\cite{arjovsky2017towards, radford2015unsupervised}, thus requiring careful initialization and proper regularization as well as other tricks~\cite{metz2016unrolled, pfau2016connecting, radford2015unsupervised, salimans2016improved}. Even so, several problems are still commonly observed including a phenomena where the generator oscillates, without ever converging to a fixed point, or mode collapse when the generator maps many latent codes $z$ to the same point, thus failing to produce diverse samples. 

The closest work related to our approach is~\cite{arora2017generalization} that showed the existence of an approximate mixed equilibrium with certain generalization properties; yet
without providing a constructive way to find such equilibria.
Instead, they advocate the use of mixed strategies, and  suggest to do so by using 
the exponentiated gradient algorithm~\cite{kivinen1997exponentiated}.
The work of~\cite{tolstikhin2017adagan} also uses a similar mixture approach based on boosting. Other works have studied the problem of equilibrium and stabilization of GANs, often relying on the use of an auto-encoder as discriminator~\cite{berthelot2017began} or jointly with the GAN models~\cite{che2016mode}. In this work, we focus on providing convergence guarantees to a \textit{mixed equilibrium} (definition in Section~\ref{sec:Semi_concaveGames}) using a technique from online optimization that  relies on the players' past actions.

\subsection{Online Learning}
Online learning is a sequential decision making framework in which a player 
aims at minimizing a cumulative loss function revealed to her
sequentially. 
The source of the loss functions may be arbitrary or even adversarial, and the player seeks to provide worst case guarantees on her performance. 
Formally, this framework can be described  as a repeated game of $T$ rounds between a player $\P_1$ and an adversary $\P_2$. At each round  $t \in[T]$: \textbf{(1)} $\P_1$ chooses a point $\u_t\in\K$ according to some algorithm $\A$, \textbf{(2)} $\P_2$  chooses a loss function $f_t\in \mathcal{F}$, \textbf{(3)} $\P_1$ suffers a loss $f_t(\u_t)$, and the loss function $f_t(\cdot)$ is revealed to her.
The adversary is usually limited to choosing losses from a structured class of objectives $\F$, most commonly linear/convex losses.
Also, the decision set $\K$ is often assumed to be convex. 
The  performance of the player's strategy is measured by the \emph{regret}, defined as,
\begin{align} \label{eq:RegretDef}
\text{Regret}_T^{\A}(f_1,\ldots,f_T)=\sum_{t=1}^T f_t(\u_t) - \min_{\u^*\in\K}\sum_{t=1}^T f_t(\u^*)~.
\end{align}
Thus, the regret measures the cumulative loss of the player compared to the loss of the \emph{best fixed decision in hindsight}.
A player aims at minimizing her regret, and we are
interested in \emph{no-regret} strategies for which players ensure an $o(T)$ regret for any loss sequence \footnote{A regret which depends linearly on $T$
is ensured by any strategy and is therefore trivial.}.

While there are several no-regret strategies, many of them may be seen as instantiations of the \emph{Follow-the-Regularized-Leader} (FTRL) algorithm where
\begin{align}\label{eq:FTRL}
\u_{t} = \arg\min_{\u\in\K} \sum_{\tau=1}^{t-1} f_\tau(\u) + \eta_t^{-1}R(\u) \qquad \textbf{(FTRL)}
\end{align}
FTRL takes the accumulated loss observed up to time $t$ and then chooses the
point in $\K$ that minimizes the accumulated loss plus a regularization term $\eta_t^{-1}R(\u)$.
The regularization term  prevents the player  from  abruptly changing her decisions between consecutive rounds\footnote{ Tikhonov regularization $\R(\u) = \|\u\|^2$ is one of the most popular regularizers.}. This property is often crucial to obtaining no-regret guarantees.
Note that FTRL is not always guaranteed to yield no-regret, and is mainly known 
to provide such guarantees in the setting where losses are linear/convex~\cite{hazan2016introduction, shalev2012online}.

\subsection{ Zero-sum Games }
Consider two players, $\P_1,\P_2$, which may choose pure decisions among continuous sets $\K_1$ and $\K_2$, respectively.
A zero-sum game is defined by a function $M: \K_1 \times \K_2 \mapsto \reals$ which sets the utilities of the players.
Concretely, upon choosing a pure strategy $(\u,\v)\in \K_1 \times \K_2$ the utility of $\P_1$ is $-M(\u,\v)$, while the utility of $\P_2$ is  $M(\u,\v)$. The goal of either $\P_1$/$\P_2$ is to maximize their worst case utilities; thus,
\begin{align}
\label{eq:MinmaxGame}
\min_{\u\in\K_1}\max_{\v\in \K_2}M(\u,\v) \quad \textbf{(Goal of $\P_1$)}, \quad
\text{ \&   } \quad
\max_{\v\in\K_2}\min_{\u\in \K_1}M(\u,\v)\quad \textbf{(Goal of $\P_2$)}
\end{align}
This definition of a game makes sense if there exists a point  $(\u^*,\v^*)$, such that neither $\P_1$ nor $\P_2$  may increase their utility by unilateral deviation.
Such a  point $(\u^*,\v^*)$ is called a \emph{Pure Nash Equilibrium}, which is formally defined as a point which satisfies the following conditions:
\begin{align*}
M(\u^*,\v^*)\leq \min_{\u\in \K_1}M(\u,\v^*), \; \text{ \&   } \; M(\u^*,\v^*)\geq \max_{\v\in \K_2}M(\u^*,\v)~.
\end{align*}
While a pure Nash equilibrium does not always exist, the pioneering work of Nash \cite{nash1950equilibrium} established that there always exists a \emph{Mixed Nash Equilibrium} (MNE or simply equilibrium), i.e., there always exist two  distributions  $\D_1, \D_2$ such that,
\begin{align*}
\E_{(\u,\v)\sim \D_1\times \D_2}[M(\u,\v)]\leq \min_{\u\in \K_1}\E_{\v\sim \D_2}[M(\u,\v)], \;
\text{\&} 
\; \E_{(\u,\v)\sim \D_1\times \D_2}[M(\u,\v)]\geq \max_{\v\in \K_2}\E_{\u\sim \D_1}[M(\u,\v)]~.
\end{align*}
Finding an exact MNE might be computationally hard, and we are usually satisfied with finding an approximate MNE. This  is  defined below,
\begin{definition}
Let $\eps>0$.
Two distributions $\D_1,\D_2$ are called $\eps$-MNE  if the following holds,
\begin{align*}
\E_{(\u,\v)\sim \D_1\times \D_2}[M(\u,\v)]&\leq \min_{\u\in \K_1}\E_{\v\sim \D_2}[M(\u,\v)]+\eps, \\
 \E_{(\u,\v)\sim \D_1\times \D_2}[M(\u,\v)]&\geq \max_{\v\in \K_2}\E_{\u\sim \D_1}[M(\u,\v)]-\eps~.
\end{align*}
\end{definition} 
\textbf{Terminology:} In the sequel when we discuss zero-sum games, we shall sometimes use the GAN terminology, relating the $\min$ player $\P_1$ as the generator, and the $\max$ player $\P_2$, as the discriminator.

\paragraph{No-Regret \& Zero-sum Games:}
In zero-sum games, no-regret algorithms may be used to find an approximate MNE. 
Unfortunately, computationally tractable no-regret algorithms do not always exist.
An exception is the setting when $M$ is convex-concave. In this case,  the players may  invoke the powerful no-regret methods from online convex optimization to (approximately) solve the game.
This seminal idea was introduced in~\cite{freund1999adaptive}, where it was demonstrated how to invoke no-regret algorithms during $T$ rounds to  obtain an approximation guarantee of  $\eps=\O(1/\sqrt{T})$ in zero-sum matrix games.
This was later improved by~\cite{daskalakis2015near,rakhlin2013optimization}, demonstrating a guarantee of $\eps=\O(\log T/T)$.
The result that we are about to present builds on the scheme of~\cite{freund1999adaptive}.


\section{Finding an Equilibrium in GANs}
\label{sec:GANs}
\textbf{Why Mixed Equilibrium?}
In this work our ultimate  goal is to efficiently find an approximate MNE for the game. 
However, in GANs, we are usually interested in designing good generators, and one might ask whether finding an equilibrium serves this cause better than  solving the minimax problem, i.e., finding $\u\in \argmin_{\u\in\K_1}\max_{\v\in\K_2}M(\u,\v)$.
Interestingly, the minimax value of a pure strategy for the generator is always higher than the minimax value of the equilibrium strategy of the generator.
The benefit of finding an equilibrium can be demonstrated on a simple zero-sum game. Consider the  following \emph{paper-rock-scissors} game, i.e. a zero-sum game with the minimax objective
$$\underset{i\in\{1,2,3\}}{\min} \underset{j\in\{1,2,3\}}{\max}M(i,j)~; \text{ where }
\tiny{
M=
\begin{bmatrix}
     0        & -1 & 1  \\
    1       &   0 & -1  \\
    -1       & 1 & 0 
\end{bmatrix}
}. $$
Solving for the minimax objective  yields a pure strategy with a minimax value of $1$; conversely, the equilibrium strategy of the $\min$ player is a uniform distribution over actions; and its minimax value  is $0$.
Thus, finding an equilibrium by allowing mixed strategies  implies a smaller minimax value (as we show in the Section~\ref{sec:minimaxVal} this is true in general).

This section  presents a method that efficiently finds an equilibrium for semi-shallow GANs  as depicted in Fig.~\ref{fig:semishallow}.
Such architectures do not induce a convex-concave game, and therefore the result of~\cite{freund1999adaptive} does not directly apply. Nevertheless, we show that semi-shallow GANs  imply an interesting game structure which gives rise to an efficient procedure for finding an equilibrium.
In Sec.~\ref{sec:Semi_shallowGans} we show that semi-shallow GANs define games with a property that we denote as semi-concave.
Later, Sec.~\ref{sec:Semi_concaveGames} provides an algorithm with provable guarantees for such games.
Finally, in Section~\ref{sec:minimaxVal}  we show that the minimax objective of the generator's equilibrium strategy is  optimal with respect to the minimax objective.

\subsection{Semi-shallow GANs}\label{sec:Semi_shallowGans}
Semi-shallow GANs do not lead to a convex-concave game.
Nonetheless, here we show that for an appropriate choice of the activation function they induce a game which is concave with respect to the discriminator. As we present in Sec.~\ref{sec:Semi_concaveGames}, this property alone allows us to efficiently find an equilibrium.

\begin{proposition}\label{prop:ConcaveGANs}
Consider the GAN objective in Eq.~\eqref{eq:GAN_objective} and assume that the adversary is a single-layer with a sigmoid activation function, meaning $h_\v(\x) = 1/(1+\exp(-\v^\top \x))$, where $\v\in \reals^n$. Then the GAN objective is concave in $\v$.
\end{proposition}
Note that the above is not restricted to the sigmoid activation function, but it also holds for other choices of activation, e.g. cumulative gaussian distribution, i.e.
$$h_\v(\x) = \Phi(\v^\top \x),\quad \text{ where } \Phi(a) = \int_{y=-\infty}^a {{(2\pi)}}^{-0.5}\exp(-y^2/2)dy~.$$
Note that the logarithm of $h_\v(x)$ for the sigmoid and cumulative gaussian activations correspond to the well known logit and probit models, \cite{mccullagh1989generalized}.

\subsection{Semi-concave Zero-sum Games}\label{sec:Semi_concaveGames}
Here we discuss the  setting of zero-sum games (see Eq.~\eqref{eq:MinmaxGame}) 
which are \textit{semi-concave}. Formally a game, $M$, is semi-concave if for any fixed  $\u_0\in\K_1$
the function $g(\v):= M(\u_0,\v)$ is concave in $\v$.
Algorithm~\ref{alg:noRegretzeroSum} presents our method for semi-concave games. This algorithm is an instantiation of the scheme derived by~\cite{freund1999adaptive}, with  specific choices of the online algorithms 
$\A_1,\A_2,$ used by the players.
Note that both $\A_1,\A_2$ are two different instances of the FTRL approach presented in Eq.~\eqref{eq:FTRL}.

Let us discuss Algorithm~\ref{alg:noRegretzeroSum} and then present its guarantees. First note that each player calculates a sequence of $T$ points based on  an online algorithm $\A_1/\A_2$.
Interestingly, the sequence of (loss/reward) functions given to the online algorithm is based on the game objective $M$, and also on the decisions made by the other player. For example, the loss sequence that $\P_1$ receives is $\{f_t(\u) := M(\u,\v_t)\}_{t\in[T]}$. 
After $T$ rounds we end up with two mixed strategies $\D_1,\D_2$, each being a uniform distribution 
over the respective online decisions $\{\u_t\}_{t\in[T]}, \{\v_t\}_{t\in[T]}$.
Note that the first decision points $\u_1,\v_1$ are set by $\A_1,\A_2$ before encountering any (loss/reward) function, and the dummy functions $f_0(\u)=0, g_0(\v)=0$ are only introduced in order to simplify the exposition. 
Since $\P_1$'s goal is to minimize, it is natural to think of the $f_t$'s as loss functions, and measure the guarantees of $\A_1$ according to the regret  as defined in Equation~\eqref{eq:RegretDef}. Analogously, since  $\P_2$'s goal is to maximize, it is natural to think of the $g_t$'s as reward functions, and measure the guarantees of $\A_2$ according to the following appropriate definition of regret,
$
\text{Regret}_T^{\A_2} 
=
 \max_{\v^*\in\K_2}\sum_{t=1}^T g_t(\v^*)-\sum_{t=1}^T g_t(\v_t)~.
$

\begin{algorithm}[t]
\caption{\FTRLGAN}
\label{alg:noRegretzeroSum}
\begin{algorithmic}
\STATE \textbf{Input}: \#steps $T$, Game objective $M(\cdot,\cdot)$ 
\FOR{$t=1 \ldots T$ }
\STATE {Calculate:} 
\begin{align*}
\text{(Alg.~$\A_1$)}\quad
\u_t\gets \text{FTRL}_1(f_0,\ldots,f_{t-1}) \quad \text{\&   }\quad
\text{(Alg.~$\A_2$)}\quad
\v_t\gets \text{FTRL}_2(g_0,\ldots,g_{t-1})
\end{align*}
\STATE {Update:}
$\qquad\qquad\qquad\quad
f_t(\cdot) = M(\cdot,\v_t) \quad \text{\&   }\quad
g_t(\cdot) = M(\u_t,\cdot) 
$
\ENDFOR
\STATE \textbf{Output mixed strategies}: $\D_1\sim \text{Uni}\{\u_1,\ldots,\u_T\}$,
$\D_2\sim \text{Uni}\{\v_1,\ldots,\v_T\}$.
\end{algorithmic}
\end{algorithm}

The following theorem presents our guarantees for semi-concave games:
\begin{theorem}\label{thm:GuaranteeSemiConcaveGame}
Let $\K_2$ be a  convex set. Also, let $M$ be  a semi-concave zero-sum game, and assume $M$ is $L$-Lipschitz countinuous.
 Then upon invoking Alg.~\ref{alg:noRegretzeroSum} for $T$ steps, using the FTRL versions  $\A_1,\A_2,$ appearing below, it outputs  mixed strategies $(\D_1,\D_2)$ that are   
$\eps$-MNE, where $\eps=O(1/\sqrt{T})$.
\begin{align*}
 \textbf{$(\A_1)$} \quad
 \u_t \gets \argmin_{\u\in \K_1} 
\sum_{\tau=0}^{t-1}f_\tau(\u)   \qquad \& \qquad
(\A_2) \quad
\v_t \gets \argmax_{\v\in \K_2} 
\sum_{\tau=0}^{t-1}\nabla g_\tau(\v_\tau)^\top \v-\frac{\sqrt{T}}{2\eta_0}\|\v\|^2 
\end{align*}
\end{theorem}
The most important point to note is that the accuracy of the approximation $\eps$ improves as the number of iterations $T$ grows. This enables obtaining arbitrarily good approximation for a large enough $T$.
Note that  $\A_1$ is in fact follow-the-leader, i.e., FTRL without regularization.
The discriminator, $\A_2$, also uses the FTRL scheme. Yet, instead of the original reward functions, $g_t(\cdot)$, it utilizes linear approximations $\tg_t(\v) = \nabla g_t(\v_t)^\top \v$. Also note the use of the (minus) square $\ell_2$ norm as regularization\footnote{Note that the minus sign in the regularization is since the discriminator's goal is to maximize, thus the $g_t(\cdot)$, may be thought of as reward functions.}.
The $\eta_0$ parameter  depends on the Lipschitz constant of $M$ as well as on the diameter
of $\K_2$ defined as, $d_2:= \max_{\v_1,\v_2\in\K_2}\|\v_1-\v_2\|$. Concretely, $\eta_0 = d_2/\sqrt{2}L $.

Next, we first provide a short proof sketch of the proof of Thm.~\ref{thm:GuaranteeSemiConcaveGame} while the full proof appears in Section~\ref{sec:analysis}.

\begin{proof}[Proof sketch]
The proof makes  use of a theorem due to~\cite{freund1999adaptive} which shows that if both $\A_1$ and $\A_2$ ensure no-regret then it implies convergence to approximate MNE.
Since the game is concave with respect to $\P_2$, it is well known that the FTRL version $\A_2$  appearing in Thm.~\ref{thm:GuaranteeSemiConcaveGame} is a no-regret strategy (see e.g.~\cite{hazan2016introduction}). The challenge  is therefore to show that $\A_1$ is also a no-regret strategy. This is non-trivial, especially for semi-concave games that do not necessarily  have any special structure with respect to the generator~\footnote{The result of~\cite{HazanKoren} shows that there does not exist any efficient no-regret algorithm, $\A_1$, in the general case where the loss sequence $\{f_t(\cdot)\}_{t\in[T]}$ received by $\A_1$ is arbitrary.}.
However, the loss sequence received by the generator is not arbitrary but rather it follows a special sequence based on the choices of the discriminator, $\{f_t(\cdot) = M(\cdot,\v_t)\}_{t}$.  In the  case of semi-concave games, the sequence of discriminator decisions, $\{\v_t\}_{t}$ has a special property which ``stabilizes" the loss sequence $\{f_t\}_{t}$, which in turn enables us to  establish no-regret for $\A_1$.
\end{proof}

\textbf{Remark:} Note that Alg.~$\A_1$ in Thm.~\ref{thm:GuaranteeSemiConcaveGame} assumes the availability of an oracle that can efficiently find  a global minimum for the FTL objective, $\sum_{\tau=0}^{t-1}f_\tau(\u)$. This involves a minimization over a sum of generative networks.
Therefore, our result may be seen as a reduction from the problem of finding an equilibrium to an offline optimization problem. This reduction is not trivial, especially in light of the negative results of \cite{HazanKoren}, which imply that in the general case finding an equilibrium is  hard, even with such an efficient offline optimization oracle at hand.
Thus, our result enables to take advantage of progress made in  supervised deep learning in order to efficiently find an equilibrium for GANs.

\subsection{Minimax value of Equilibrium Strategy}\label{sec:minimaxVal}
In GANs we are mainly interested in ensuring the performance of the generator (resp. discriminator)
with respect to the \emph{minimax} (resp. \emph{maximin}) objective. 
Let $(\D_1,\D_2)$ the pair of mixed strategies that Algorithm~\ref{alg:noRegretzeroSum} outputs.
Note that the minimax value of  $\D_1$ might be considerably smaller than the pure minimax value,  as is shown in  the example regarding the
 \emph{paper-rock-scissors} game (see Sec.~\ref{sec:GANs}).
The next lemma shows that the mixed strategy $\D_1$ is always  (approximately) better with respect to the pure minimax value,(see proof in appendix~\ref{sec:proof_lemmMinmaxval})
\begin{lemma}\label{lem:NoRegretGamesMinMax}
The mixed strategy $\D_1$ that Algorithm~\ref{alg:noRegretzeroSum} outputs is $\eps$-optimal with respect to the minimax value, i.e.,
$$\max_{\v\in\K_2} \E_{\u\sim\D_1}[M(\u,\v)]\leq \min_{\u\in\K_1}\max_{\v\in\K_2}M(\u,\v)+\eps$$
where $\eps$ here is equal to the one defined in Thm.~\ref{thm:NoRegretGames}.
\end{lemma} 
Analogous result hold for $\D_2$ with respect to the pure maximin objective.

\subsection{Practical \FTRLGAN Algorithm}
\begin{algorithm}[h]
\caption{Practical \FTRLGAN}
\label{alg:noRegretzeroSumPractice}
\begin{algorithmic}
\STATE \textbf{Input}: \#steps $T$, Game objective $M(\cdot,\cdot)$, number of past states $K$, spacing $m$
\STATE \textbf{Initialize}: Set loss/reward $f_0(\cdot)=0,g_0(\cdot)=0$, initialize queues $\Q_1$.insert($f_0$), $\Q_2$.insert($g_0$)
\FOR{$t=1 \ldots T$ }
\STATE Update generator and discriminator based on a minibatch of noise samples and data samples:
{\small
\begin{align*}
\u_{t+1} \gets\u_t- \nabla_{\u_t}\left( \frac{1}{|\Q_1|}\sum_{f\in\Q_1}f(\u) +\frac{C}{\sqrt{t}}\|\u\|^2\right)
\; \text{\&   }\;
\v_{t+1} \gets\v_t-  \nabla_{\v_t}\left( \frac{1}{|\Q_2|}\sum_{g\in\Q_2}g(\v) -\frac{C}{\sqrt{t}}\|\v\|^2\right)
\end{align*}
}
\STATE {Calculate:}
$
f_t(\cdot) = M(\cdot,\v_t) \; \text{\&   }\;
g_t(\cdot) = M(\u_t,\cdot) 
$ 
\\
\STATE Update $\Q_1$ and $\Q_2$ (see main text or Algorithm~\ref{alg:A1_A2} in the appendix)
\ENDFOR
\end{algorithmic}
\end{algorithm}

\label{sec:practical_algorithm}
In this section we describe a practical application of the general approach described in Alg.~\ref{alg:noRegretzeroSum} to common \emph{deep} GAN architectures.
There are several  differences compared to the theoretical algorithm that we have presented:
\textbf{(i)} We use the FTRL objective (Eq~\eqref{eq:FTRL}) for both players. Note that Alg.~$\A_2$ appearing in Thm~\ref{thm:GuaranteeSemiConcaveGame} uses FTRL with linear approximations, which is only appropriate for semi-concave games.  
\textbf{(ii)} As calculating the global minimizer of the FTRL objective is impractical, we update the weights based  on the gradients of the FTRL objective, using traditional optimization techniques such as SGD or Adam. This differs from the standard GAN training which only employs the gradient of the last loss/reward function.
\textbf{(iii)} The full FTRL algorithm requires to save the entire history of past generators/discriminators, which is computationally intractable. We find it sufficient to maintain a summary of the history, using a small number of representatives. In order to capture a diverse subset of the history, we keep a queue $\Q$ containing $K:=|\Q|$ models whose spacing between each other is determined by the following heuristic. Every $m $ update steps, we remove the oldest model in the queue and add the current one. The number of steps between switches, $m$,  can be set as a constant, but we find it more effective to keep it small at the beginning and increase its value as the number of rounds increases. We hypothesize that as the training progresses and the individual models become more powerful, we should switch the models at a lower rate, keeping them more spaced out. The pseudo-code and a detailed description of the algorithm appears in the Appendix.


\section{Experimental results}
\label{sec:experiments}

We now compare \FTRLGAN to various baselines and demonstrate improved stability and sample diversity. We test our method on models where the traditional GAN training has difficulties converging and engages in a behavior of mode collapse. We also perform experiments on harder tasks using the DCGAN architecture~\cite{radford2015unsupervised} for which we show that \FTRLGAN reduces mode dropping while retaining high visual sample quality. For all of the experiments, we generate from the newest generator only. Experimental details and comparisons to additional baselines, as well as a set of recommended hyperparameters are available in Appendix~\ref{sec:app_experiments} and Appendix~\ref{sec:practical_GAN}, respectively. 

\subsection{Non-convergence and Mode Dropping}

\subsubsection{Toy Dataset: Mixture of Gaussians}

\begin{figure*}[ht]
	\begin{center}
          \begin{tabular}{@{}S@{\hspace{2mm}}T@{\hspace{15mm}}T}
          & & \small{\textsc{Target}} \\
          \textsc{GAN} & \includegraphics[width=13cm,height=1.5cm,keepaspectratio]{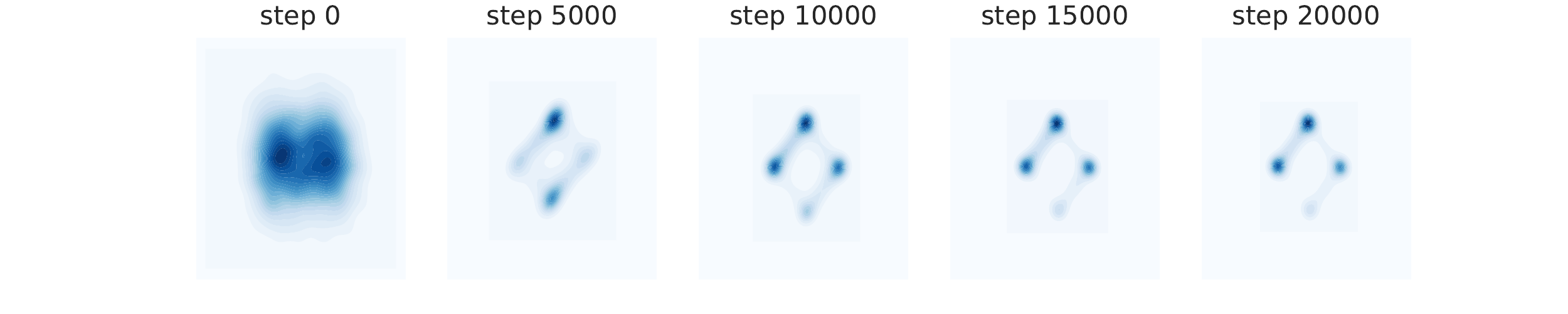}
&
		\includegraphics[width=1.5cm,height=1.3cm,keepaspectratio]{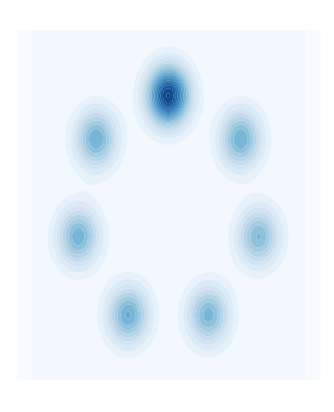} \\
          \textbf{\FTRLGAN} & \includegraphics[width=13cm,height=1.5cm,keepaspectratio]{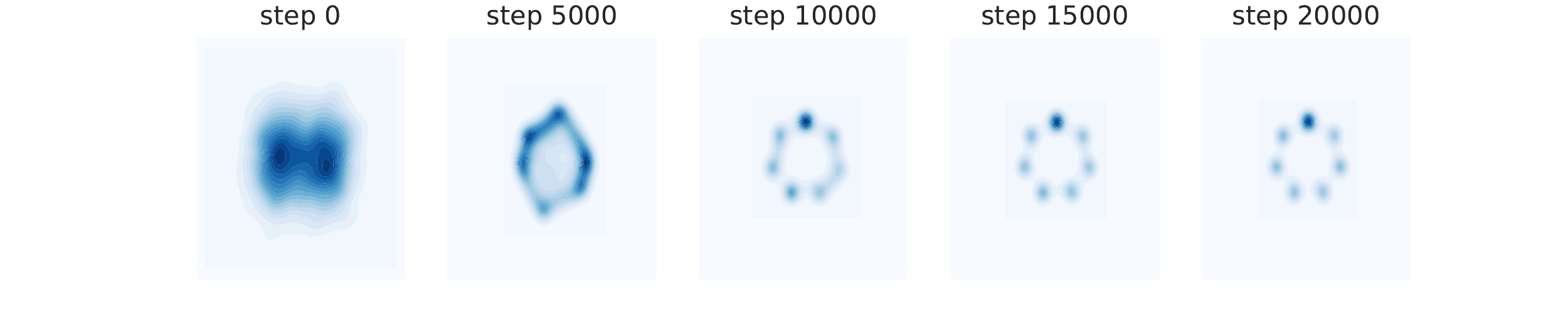}
&
		\includegraphics[width=1.5cm,height=1.3cm,keepaspectratio]{Images/ToyExperiment/target.png} 
	  \end{tabular}
  \caption{\small{Mode Collapse on a Gaussian Mixture. We show heat maps of the generator distribution over time, as well as the target data distribution in the last column. Standard GAN updates (top row) cause mode collapse, whereas \FTRLGAN using $K=5$ past steps (bottom row) spreads its mass over all the modes of the target distribution.}}
  \label{Experiments:Mixture}
	\end{center}
\end{figure*}

We first train a simple architecture using the standard GAN approach as well as \FTRLGAN on a synthesized 2D dataset following~\cite{metz2016unrolled, che2016mode}. The data consists of a mixture of 7 Gaussians whose centers are aligned in a circle. On this dataset, it can directly be seen how the traditional GAN updates lead to mode dropping where one observes a collapse of large volumes of probability mass onto a few modes. This is hypothesised to be due to the differences of the minimax and maximin solutions of the game~\cite{goodfellow2016nips}. If the order of the min and max operations switch, the minimization with respect to the generator's parameters is performed in the inner loop. This causes the generator to map every latent code to one or very few points for which the discriminator believes are likely. As simultaneous gradient descent updates do not clearly prioritize any specific ordering of minimax or maximin, in practice we often obtain results that resemble the latter. This phenomena is clearly observed in Fig.~\ref{Experiments:Mixture}. 
In contrast, \FTRLGAN takes advantage of the history of the player's actions which yields better gradient information. Intuitively, the generator is updated such that it fools the past discriminators. In order to do so, the generator has to spread its mass more fairly according to the true data distribution.

\subsubsection{Augmented MNIST}

We now evaluate the ability of our approach to avoid mode collapse on real image data coming from an augmented version of the MNIST dataset. Similarly to~\cite{metz2016unrolled, che2016mode}, we combine three randomly selected MNIST digits to form 3-channel images, resulting in a dataset with 1000 different classes, one for each of the possible combinations of the ten MNIST digits.

We train a simplified DCGAN architecture (see details in Appendix~\ref{sec:app_experiments}) with both GAN and \FTRLGAN with a different number of saved past states. The evaluation of each model is done as follows. We generate a fixed amount of samples (25,600) from each model and classify them using a pre-trained MNIST classifier with an accuracy of $~99.99\%$. The models that exhibit less mode collapse are expected to generate samples from most of the 1000 modes.

We report two different evaluation metrics in Table~\ref{tbl:ResultsMNIST}: i) the number of classes for which a model generated at least one sample, and ii) the reverse KL divergence. The reverse KL divergence between the model and the target data distribution is computed by considering that the data distribution is a uniform distribution over all classes.

\begin{table}[ht]
  \begin{center}
  \begin{tabular}{@{}|c|c|c|c|@{}}
    \hline
     Models & 0 states (GAN) & 5 states & 10 states \\ \hline
      Generated Classes & 629 $\pm$ 121.08  & 743 $\pm$ 64.31 &  795 $\pm$ 37  \\\hline
      Reverse KL & 1.96 $\pm$ 0.64 &  1.40 $\pm$ 0.21 &  1.24 $\pm$ 0.17\\\hline
  \end{tabular}
  \caption{\small{Stacked MNIST: Number of generated classes out of 1000 possible combinations, and the reverse KL divergence score. The results are averaged over 10 runs. }}
  \label{tbl:ResultsMNIST}
  \end{center}
\end{table}

\subsection{Image Modeling}

We turn to the evaluation of our model for the task of generating rich image data for which the modes of the data distribution are unknown. In the following, we perform experiments that indirectly measure mode coverage through metrics based on the sample diversity and quality.

\subsubsection{Inference via Optimization on CIFAR10}

We train a DCGAN architecture on CIFAR10~\cite{krizhevsky2009learning} and evaluate the performance of each model using the inference via optimization technique introduced in~\cite{metz2016unrolled} and explained in Appendix~\ref{sec:inf_optimization}.

The average MSE over 10 rounds using different seeds is reported in Table \ref{tbl:Cifar10Scores}. Using \FTRLGAN with as few as 5 past states results in a significant gain which can be further improved by increasing the number of past states to 10 and 25. In addition, the training procedure becomes more stable as indicated by the decrease in the standard deviation. The percentage of minibatches that achieve the lowest reconstruction loss with the different models is given in Table~\ref{tbl:Cifar10Scores}. This can also be visualized by comparing the closest images $x_{closest}$ from each model to real target images $x_{target}$ as shown in Figure~\ref{tbl:VisualCifar}. The images are randomly selected images from the batch which has the largest absolute difference in MSE between GAN and \FTRLGAN with 25 states. The samples obtained by the original GAN are often blurry while samples from \FTRLGAN are both sharper and exhibit more variety, suggesting a better coverage of the true data distribution.

\begin{table}[ht]
\centering
\begin{tabular}{|c|c|c|c|c|c|}
\hline
Target                                                               & Past States    & 0 (GAN)        & 5 states         & 10 states        & 25 states                 \\ \hline
\multirow{2}{*}{\begin{tabular}[c]{@{}c@{}}Train\\ Set\end{tabular}} & MSE            & 61.13 $\pm$ 3.99 & 58.84 $\pm$ 3.67 & 56.99 $\pm$ 3.49 & \textbf{48.42 $\pm$ 2.99} \\ \cline{2-6} 
                                                                     & Best Rank (\%) & 0 \%                & 0  \%              & 18.66 \%         & \textbf{81.33 \%}         \\ \hline
\multirow{2}{*}{\begin{tabular}[c]{@{}c@{}}Test\\ Set\end{tabular}}  & MSE            & 59.5 $\pm$ 3.65  & 56.66 $\pm$ 3.60 & 53.75 $\pm$ 3.47 & \textbf{46.82 $\pm$ 2.96} \\ \cline{2-6} 
                                                                     & Best Rank (\%) & 0 \%                & 0 \%                & 17.57 \%         & \textbf{82.43 \%}         \\ \hline
\end{tabular}
\caption{\small{CIFAR10: MSE between target images from the train and test set and the best rank which consists of the percentage of minibatches containing  target images that can be reconstructed with the lowest loss across various models. We use 20 different minibatches, each containing 64 target images. Increasing the number of past states for \FTRLGAN allows the model to better match the real images.}}
\label{tbl:Cifar10Scores}
\end{table}

\FloatBarrier

\begin{table}[ht]
\setlength\tabcolsep{1.5pt}

  \begin{minipage}{0.48\textwidth}  
  \begin{tabular}{@{}c|ccc@{}}
    \hline
    Real Image & 0 states  & 10 states & 25 states\\ \hline
    
      & 4.14 & 2.37 & 2.19  \\
      \includegraphics[width=1.0cm,height=1.0cm,keepaspectratio]{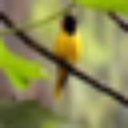}
      &
      \includegraphics[width=1.0cm,height=1.0cm,keepaspectratio]{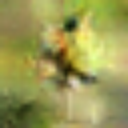}
      &
      \includegraphics[width=1.0cm,height=1.0cm,keepaspectratio]{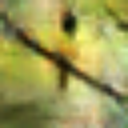}
      &
      \includegraphics[width=1.0cm,height=1.0cm,keepaspectratio]{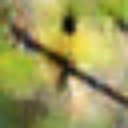}
\\ \hline
& 4.17 & 2.97 & 2.58  \\
      \includegraphics[width=1.0cm,height=1.0cm,keepaspectratio]{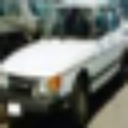}
      &
      \includegraphics[width=1.0cm,height=1.0cm,keepaspectratio]{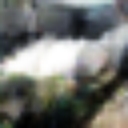}
      &
      \includegraphics[width=1.0cm,height=1.0cm,keepaspectratio]{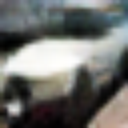}
      &
      \includegraphics[width=1.0cm,height=1.0cm,keepaspectratio]{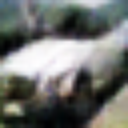}
\\ \hline
\end{tabular}%
\end{minipage}
\begin{minipage}{0.48\textwidth}  
\begin{tabular}{@{}c|ccc@{}}
    \hline
    Real Image & 0 states& 10 states & 25 states\\ \hline
    
      & 3.06 & 2.51 & 2.35  \\
      \includegraphics[width=1.0cm,height=1.0cm,keepaspectratio]{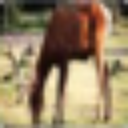}
      &
      \includegraphics[width=1.0cm,height=1.0cm,keepaspectratio]{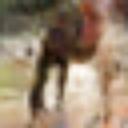}
      &
      \includegraphics[width=1.0cm,height=1.0cm,keepaspectratio]{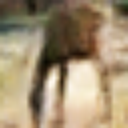}
      &
      \includegraphics[width=1.0cm,height=1.0cm,keepaspectratio]{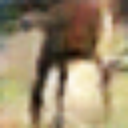}
\\ \hline
& 1.12 & 0.30 & 0.39  \\
      \includegraphics[width=1.0cm,height=1.0cm,keepaspectratio]{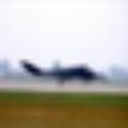}
      &
      \includegraphics[width=1.0cm,height=1.0cm,keepaspectratio]{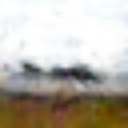}
      &
      \includegraphics[width=1.0cm,height=1.0cm,keepaspectratio]{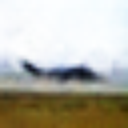}
      &
      \includegraphics[width=1.0cm,height=1.0cm,keepaspectratio]{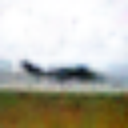}
\\ \hline
\end{tabular}
\end{minipage}
 \caption{\small{CIFAR10: Target images from the test set are shown on the left. The images from each model that best resemble the target image are shown for different number of past states: 0 (GAN), 10 and 25 (\FTRLGAN). The reconstruction MSE loss is indicated above each image.}}
\label{tbl:VisualCifar}
\end{table}
\FloatBarrier

\subsubsection{Estimation of Missing Modes on CelebA}

We estimate the number of missing modes on the CelebA dataset~\cite{liu2015deep} by using an auxiliary discriminator as performed in~\cite{che2016mode}. The experiment consists of two phases. In the first phase we train GAN and \FTRLGAN models and generate a fixed number of images. In the second phase we independently train a noisy discriminator using the DCGAN architecture where the training data is the previously generated data from each of the models, respectively. The noisy discriminator is then used as a mode estimator. Test images from CelebA are provided to the mode estimator and the number of images that are classified as fake can be viewed as images on a missing mode. Table~\ref{tbl:ResultsCelebA} showcases number of missed modes for the two models. Generated samples from each model are given in the Appendix.

\begin{table}[ht]
  \centering
  \begin{tabular}{@{}|c|c|c|@{}}
    \hline
    $\sigma$ & 0 states (GAN)  & 5 states (\FTRLGAN) \\ \hline
      0.25 & 3004 $\pm$ 4154 & 1407 $\pm$ 1848  \\\hline
      0.5 & 2568.25 $\pm$ 4148 & 1007 $\pm$ 1805 \\\hline
  \end{tabular}
  \caption{\small{CelebA: Number of images from the test set that the auxiliary discriminator classifies as not real. Gaussian noise with variance $\sigma^2$ is added to the input of the auxiliary discriminator, with the standard deviation shown in the first row. The test set consists of 50,000 images.}}
  \label{tbl:ResultsCelebA}
\end{table}

Interestingly, even with small number of past states (K=5), \FTRLGAN manages to stabilize the training and generate more diverse samples on all the datasets. In terms of computational complexity, our algorithm scales linearly with $K$. However, all the elements in the sum are independent and can be computed efficiently in a parallel manner.


\section{Analysis}
\label{sec:analysis}
Here we provide the proof of Thm.~\ref{thm:GuaranteeSemiConcaveGame}.

\begin{proof}
We make a use of a theorem due to~\cite{freund1999adaptive} which shows that if both $\A_1,\A_2$ ensure no regret implies approximate MNE. For completeness we provide its proof in Sec.~\ref{sec:Nash}.
\begin{theorem}\label{thm:NoRegretGames}
The mixed strategies $(\D_1,\D_2)$ that Algorithm~\ref{alg:noRegretzeroSum} outputs are $\eps$-MNE, where
$$\eps : =  \left({B^{\A_1}_T}+ {B^{\A_2}_T}\right)/{T}~.$$
here $B^{\A_1}_T,B^{\A_2}_T$ are bounds on the regret of $\A_1,\A_2$.
\end{theorem} 
According to Thm.~\ref{thm:NoRegretGames}, it is sufficient to show that both $\A_1$, and $\A_2$ ensure a regret bound of $O(\sqrt{T})$.

\textbf{Guarantees for $\A_2$:}
this FTRL version  is  well known  in online learning, and its regret guarantees can be found in the literature, (e.g, Theorem 5.1 in ~\cite{hazan2016introduction}). The following lemma provides its guarantees,
\begin{lemma}\label{lem:ConcaveFTRL_guarantees}
Let $d_2$ be the diameter of $\K_2$. 
 Invoking  $\A_2$  with $\eta_0 = d_2/\sqrt{2}L$,
ensures the following regret bound over the sequence of concave functions $\{g_t\}_{t\in[T]}$,
\begin{align*} 
\text{Regret}_T^{\A_2}(g_1,\ldots,g_T)\leq Ld_2\sqrt{2T}~.
\end{align*}
Moreover, the following applies for the sequence $\{\v_t\}_{t\in[T]}$ generated by $\A_2$,
\begin{align*} 
\|\v_{t+1}-\v_t\|\leq d_2/\sqrt{2T}~.
\end{align*}
\end{lemma}
Note that the proof heavily relies on the concavity of the $g_t(\cdot)$'s, which is due to the concavity of the game with respect to the discriminator. For completeness we provide a proof of 
the second part of the lemma in Sec.~\ref{sec:SecondPart}.

\textbf{Guarantees for $\A_1$:}
By Lemma~\ref{lem:ConcaveFTRL_guarantees}, the sequence generated by the discriminator not only ensures low regret but is also stable in the sense that consecutive decision points are close by. This is the key property which will enable us to establish a regret bound for algorithm $\A_1$.
Next we  state the guarantees of $\A_1$,

\begin{lemma}\label{lem:FTL}
Let $C := \max_{\u\in\K_1,\v\in\K_2} |M(\u,\v)|$. Consider the loss sequence appearing in Alg~\ref{alg:noRegretzeroSum}, $\{f_t(\cdot): = M(\cdot,\v_t)\}_{t\in [T]}$.
Then algorithm $\A_1$ ensures the following regret bound over this sequence,
\begin{align*} 
\text{Regret}_T^{\A_1}(f_1,\ldots,f_T)\leq \frac{1}{\sqrt{2}}Ld_2\sqrt{T}+2C~.
\end{align*}
\end{lemma}
Combining the regret bounds of Lemmas~\ref{lem:ConcaveFTRL_guarantees},~\ref{lem:FTL} and Theorem~\ref{thm:NoRegretGames} concludes the proof of Thm.~\ref{thm:GuaranteeSemiConcaveGame}.
\end{proof}

\subsection{Proof of Lemma~\ref{lem:FTL}}
\begin{proof}
We use the following regret bound regarding the FTL (follow-the-leader) decision rule, derived in~\cite{kalai2005efficient} (see also \cite{shalev2012online}),
\begin{lemma}
For any sequence of loss functions $\{f_t\}_{t\in[T]}$, the regret of  FTL  is bounded as follows,
\begin{align*} 
\text{Regret}_T^{\text{FTL}}(f_1,\ldots,f_T)\leq \sum_{t=1}^T \blr{f_t(\u_t)-f_t(\u_{t+1})}~.
\end{align*}
\end{lemma}
Since $\A_1$ is FTL, the above bound applies. Thus, using the above bound together with  the stability of the $\{\v_t\}_{t\in[T]}$ sequence we obtain,
\begin{align*} 
\text{Regret}_T^{\A_1}
&\leq
 \sum_{t=1}^T \blr{f_t(\u_t)-f_t(\u_{t+1})} \\
&
= \sum_{t=1}^{T-1} \blr{f_t(\u_t)-f_t(\u_{t+1}) + f_{t+1}(\u_{t+1}) - f_{t+1}(\u_{t+1})} + \blr{ f_T(\u_T)-f_T(\u_{T+1})} \\
&=
 \sum_{t=1}^{T-1} \blr{ f_{t+1}(\u_{t+1})-f_t(\u_{t+1})} + \sum_{t=1}^{T-1} \blr{f_t(\u_t) - f_{t+1}(\u_{t+1})}
+ \blr{ f_T(\u_T)-f_T(\u_{T+1})} \\
&=
 \sum_{t=1}^{T-1}\blr{ M(\u_{t+1},\v_{t+1}) - M(\u_{t+1},\v_t)} +\blr{ f_1(\u_1) - f_{T}(\u_{T+1})}\\
&\leq
\sum_{t=1}^{T-1}L\|\v_{t+1} - \v_t\| +\blr{ f_1(\u_1) - f_{T}(\u_{T+1})}\\
&\leq
LTd_2/\sqrt{2T}   +\blr{ f_1(\u_1) - f_{T}(\u_{T+1})} \\
&\leq L d_2 \sqrt{T}/\sqrt{2} + 2C~.
\end{align*}
where the fourth line uses $f_t(\cdot):=M(\cdot,\v_t)$, the fifth line uses the Lipschitz continuity of $M$. And the sixth line used the stability of the $\v_t$'s due to Lemma~\ref{lem:ConcaveFTRL_guarantees}. Finally, we use $|f_t(u)|=|M(u,v_t)|\leq C$.
\end{proof}

\subsection{Proof of Theorem~\ref{thm:NoRegretGames}}
\label{sec:Nash}
\begin{proof}
Writing explicitly $f_t(\u):= M(\u,\v_t)$ and $g_t(\v):= M(\u_t,\v)$, and plugging these into the regret guarantees of $\A_1,\A_2$, we have,
\begin{align}
&\sum_{t=1}^T M(\u_t,\v_t)-\min_{\u\in \K_1}\sum_{t=1}^T M(\u,\v_t) \leq B^{\A_1}_T~, \label{eq:Reg1}\\
&\sum_{t=1}^T -M(\u_t,\v_t)-\min_{\v\in \K_2}\sum_{t=1}^T -M(\u_t,\v) \leq B^{\A_2}_T  \label{eq:Reg2}~.
\end{align}
By definition, $\min_{\u\in \K_1}\sum_{t=1}^T M(\u,\v_t) \leq \E_{\u\sim\D_1}[\sum_{t=1}^T M(\u,\v_t)]$. Using this together with  Equation~\eqref{eq:Reg1}, we get,
\begin{align}\label{eq:Reg11}
&\sum_{t=1}^T M(\u_t,\v_t)-\E_{\u\sim\D_1}[\sum_{t=1}^T M(\u,\v_t)] \leq B^{\A_1}_T~, 
\end{align}
Summing Equations~\eqref{eq:Reg2},\eqref{eq:Reg11}, and dividing by $T$, we get,
\begin{align} \label{eq:Reg_rel}
\max_{\v\in\K_2}\frac{1}{T}\sum_{t=1}^T M(\u_t,\v)-\E_{\u\sim\D_1}[\frac{1}{T}\sum_{t=1}^T M(\u,\v_t)] \leq  \frac{B^{\A_1}_T}{T} + \frac{B^{\A_2}_T}{T}~. 
\end{align}
Recalling that $\D_1\sim\text{Uni}\{\u_1,\ldots,\u_t\},\D_2\sim\text{Uni}\{\v_1,\ldots,\v_t\}$, and denoting $\eps: = \frac{B^{\A_1}_T}{T} + \frac{B^{\A_2}_T}{T}$, we conclude that,
\begin{align*} 
\E_{(\u,\v)\sim\D_1\times \D_2}[ M(\u,\v)] \geq  \max_{\v\in\K_2}\E_{\u\sim\D_1}[M(\u,\v)]-\eps~. 
\end{align*}
We can similarly show that,
\begin{align*} 
\E_{(\u,\v)\sim\D_1\times \D_2}[ M(\u,\v)] \leq  \min_{\u\in\K_1}\E_{\v\sim\D_2}[M(\u,\v)]+\eps~. 
\end{align*}
which concludes the proof.
 \end{proof}


\section{Conclusion}
\label{sec:conclusion}
\vspace{-2mm}
We have presented a principled approach to training GANs,
which is guaranteed to reach convergence to a mixed equilibrium for semi-shallow architectures. Empirically, our approach presents several advantages when applied to commonly used GAN architectures, such as improved stability or reduction in mode dropping. Our results open an avenue for the use of online-learning and game-theoretic techniques in the context of training GANs. One question that remains open is whether the theoretical guarantees can be extended to more complex architectures.

\section*{Acknowledgement}
The authors would like to thank Hoda Heidari and  Johannes Kirschner for helpful comments and suggestions. 
This research was partially supported by ERC StG 307036.
This work was done in part while Andreas Krause was visiting the Simons Institute for the Theory of Computing.
K.Y.L. is supported by the ETH Z\"urich Postdoctoral Fellowship and Marie Curie Actions for People COFUND program.

\bibliographystyle{abbrvnat}
\bibliography{bib}

\begin{thebibliography}{31}
\providecommand{\natexlab}[1]{#1}
\providecommand{\url}[1]{\texttt{#1}}
\expandafter\ifx\csname urlstyle\endcsname\relax
  \providecommand{\doi}[1]{doi: #1}\else
  \providecommand{\doi}{doi: \begingroup \urlstyle{rm}\Url}\fi

\bibitem[Arjovsky and Bottou(2017)]{arjovsky2017towards}
M.~Arjovsky and L.~Bottou.
\newblock Towards principled methods for training generative adversarial
  networks.
\newblock In \emph{NIPS 2016 Workshop on Adversarial Training. In review for
  ICLR}, volume 2016, 2017.

\bibitem[Arora et~al.(2017)Arora, Ge, Liang, Ma, and
  Zhang]{arora2017generalization}
S.~Arora, R.~Ge, Y.~Liang, T.~Ma, and Y.~Zhang.
\newblock Generalization and equilibrium in generative adversarial nets (gans).
\newblock \emph{arXiv preprint arXiv:1703.00573}, 2017.

\bibitem[Berthelot et~al.(2017)Berthelot, Schumm, and Metz]{berthelot2017began}
D.~Berthelot, T.~Schumm, and L.~Metz.
\newblock Began: Boundary equilibrium generative adversarial networks.
\newblock \emph{arXiv preprint arXiv:1703.10717}, 2017.

\bibitem[Che et~al.(2016)Che, Li, Jacob, Bengio, and Li]{che2016mode}
T.~Che, Y.~Li, A.~P. Jacob, Y.~Bengio, and W.~Li.
\newblock Mode regularized generative adversarial networks.
\newblock \emph{arXiv preprint arXiv:1612.02136}, 2016.

\bibitem[Daskalakis et~al.(2015)Daskalakis, Deckelbaum, and
  Kim]{daskalakis2015near}
C.~Daskalakis, A.~Deckelbaum, and A.~Kim.
\newblock Near-optimal no-regret algorithms for zero-sum games.
\newblock \emph{Games and Economic Behavior}, 92:\penalty0 327--348, 2015.

\bibitem[Freund and Schapire(1999)]{freund1999adaptive}
Y.~Freund and R.~E. Schapire.
\newblock Adaptive game playing using multiplicative weights.
\newblock \emph{Games and Economic Behavior}, 29\penalty0 (1-2):\penalty0
  79--103, 1999.

\bibitem[Glorot and Bengio(2010)]{glorot2010understanding}
X.~Glorot and Y.~Bengio.
\newblock Understanding the difficulty of training deep feedforward neural
  networks.
\newblock In \emph{Aistats}, volume~9, pages 249--256, 2010.

\bibitem[Goodfellow(2016)]{goodfellow2016nips}
I.~Goodfellow.
\newblock Nips 2016 tutorial: Generative adversarial networks.
\newblock \emph{arXiv preprint arXiv:1701.00160}, 2016.

\bibitem[Goodfellow et~al.(2014)Goodfellow, Pouget-Abadie, Mirza, Xu,
  Warde-Farley, Ozair, Courville, and Bengio]{Goodfellow:2014td}
I.~Goodfellow, J.~Pouget-Abadie, M.~Mirza, B.~Xu, D.~Warde-Farley, S.~Ozair,
  A.~Courville, and Y.~Bengio.
\newblock {Generative Adversarial Nets}.
\newblock pages 2672--2680, 2014.

\bibitem[Gretton et~al.(2012)Gretton, Borgwardt, Rasch, Sch{\"o}lkopf, and
  Smola]{gretton2012kernel}
A.~Gretton, K.~M. Borgwardt, M.~J. Rasch, B.~Sch{\"o}lkopf, and A.~Smola.
\newblock A kernel two-sample test.
\newblock \emph{Journal of Machine Learning Research}, 13\penalty0
  (Mar):\penalty0 723--773, 2012.

\bibitem[Hazan and Koren(2016)]{HazanKoren}
E.~Hazan and T.~Koren.
\newblock The computational power of optimization in online learning.
\newblock In \emph{Proc.\ STOC}, pages 128--141. ACM, 2016.

\bibitem[Hazan et~al.(2016)]{hazan2016introduction}
E.~Hazan et~al.
\newblock Introduction to online convex optimization.
\newblock \emph{Foundations and Trends{\textregistered} in Optimization},
  2\penalty0 (3-4):\penalty0 157--325, 2016.

\bibitem[Kalai and Vempala(2005)]{kalai2005efficient}
A.~Kalai and S.~Vempala.
\newblock Efficient algorithms for online decision problems.
\newblock \emph{Journal of Computer and System Sciences}, 71\penalty0
  (3):\penalty0 291--307, 2005.

\bibitem[Kingma and Ba(2014)]{kingma2014adam}
D.~Kingma and J.~Ba.
\newblock Adam: A method for stochastic optimization.
\newblock \emph{arXiv preprint arXiv:1412.6980}, 2014.

\bibitem[Kingma and Welling(2013)]{Kingma:2013tz}
D.~P. Kingma and M.~Welling.
\newblock {Auto-Encoding Variational Bayes}.
\newblock \emph{arXiv.org}, Dec. 2013.

\bibitem[Kivinen and Warmuth(1997)]{kivinen1997exponentiated}
J.~Kivinen and M.~K. Warmuth.
\newblock Exponentiated gradient versus gradient descent for linear predictors.
\newblock \emph{Information and Computation}, 132\penalty0 (1):\penalty0 1--63,
  1997.

\bibitem[Krizhevsky and Hinton(2009)]{krizhevsky2009learning}
A.~Krizhevsky and G.~Hinton.
\newblock Learning multiple layers of features from tiny images.
\newblock 2009.

\bibitem[Li et~al.(2015)Li, Swersky, and Zemel]{li2015generative}
Y.~Li, K.~Swersky, and R.~S. Zemel.
\newblock Generative moment matching networks.
\newblock In \emph{ICML}, pages 1718--1727, 2015.

\bibitem[Liu et~al.(2015)Liu, Luo, Wang, and Tang]{liu2015deep}
Z.~Liu, P.~Luo, X.~Wang, and X.~Tang.
\newblock Deep learning face attributes in the wild.
\newblock In \emph{Proceedings of the IEEE International Conference on Computer
  Vision}, pages 3730--3738, 2015.

\bibitem[McCullagh and Nelder(1989)]{mccullagh1989generalized}
P.~McCullagh and J.~A. Nelder.
\newblock Generalized linear models, no. 37 in monograph on statistics and
  applied probability, 1989.

\bibitem[Metz et~al.(2016)Metz, Poole, Pfau, and
  Sohl-Dickstein]{metz2016unrolled}
L.~Metz, B.~Poole, D.~Pfau, and J.~Sohl-Dickstein.
\newblock Unrolled generative adversarial networks.
\newblock \emph{arXiv preprint arXiv:1611.02163}, 2016.

\bibitem[M{\"u}ller(1997)]{muller1997integral}
A.~M{\"u}ller.
\newblock Integral probability metrics and their generating classes of
  functions.
\newblock \emph{Advances in Applied Probability}, 29\penalty0 (02):\penalty0
  429--443, 1997.

\bibitem[Nash et~al.(1950)]{nash1950equilibrium}
J.~F. Nash et~al.
\newblock Equilibrium points in n-person games.
\newblock \emph{Proceedings of the national academy of sciences}, 36\penalty0
  (1):\penalty0 48--49, 1950.

\bibitem[Nowozin et~al.(2016)Nowozin, Cseke, and Tomioka]{nowozin2016f}
S.~Nowozin, B.~Cseke, and R.~Tomioka.
\newblock f-gan: Training generative neural samplers using variational
  divergence minimization.
\newblock In \emph{Advances in Neural Information Processing Systems}, pages
  271--279, 2016.

\bibitem[Pfau and Vinyals(2016)]{pfau2016connecting}
D.~Pfau and O.~Vinyals.
\newblock Connecting generative adversarial networks and actor-critic methods.
\newblock \emph{arXiv preprint arXiv:1610.01945}, 2016.

\bibitem[Radford et~al.(2015)Radford, Metz, and
  Chintala]{radford2015unsupervised}
A.~Radford, L.~Metz, and S.~Chintala.
\newblock Unsupervised representation learning with deep convolutional
  generative adversarial networks.
\newblock \emph{arXiv preprint arXiv:1511.06434}, 2015.

\bibitem[Rakhlin and Sridharan(2013)]{rakhlin2013optimization}
S.~Rakhlin and K.~Sridharan.
\newblock Optimization, learning, and games with predictable sequences.
\newblock In \emph{Advances in Neural Information Processing Systems}, pages
  3066--3074, 2013.

\bibitem[Rezende et~al.(2014)Rezende, Mohamed, and Wierstra]{Rezende:2014vm}
D.~J. Rezende, S.~Mohamed, and D.~Wierstra.
\newblock {Stochastic backpropagation and approximate inference in deep
  generative models}.
\newblock \emph{arXiv.org}, 2014.

\bibitem[Salimans et~al.(2016)Salimans, Goodfellow, Zaremba, Cheung, Radford,
  and Chen]{salimans2016improved}
T.~Salimans, I.~Goodfellow, W.~Zaremba, V.~Cheung, A.~Radford, and X.~Chen.
\newblock Improved techniques for training gans.
\newblock In \emph{Advances in Neural Information Processing Systems}, pages
  2226--2234, 2016.

\bibitem[Shalev-Shwartz et~al.(2012)]{shalev2012online}
S.~Shalev-Shwartz et~al.
\newblock Online learning and online convex optimization.
\newblock \emph{Foundations and Trends{\textregistered} in Machine Learning},
  4\penalty0 (2):\penalty0 107--194, 2012.

\bibitem[Tolstikhin et~al.(2017)Tolstikhin, Gelly, Bousquet, Simon-Gabriel, and
  Sch{\"o}lkopf]{tolstikhin2017adagan}
I.~Tolstikhin, S.~Gelly, O.~Bousquet, C.-J. Simon-Gabriel, and
  B.~Sch{\"o}lkopf.
\newblock Adagan: Boosting generative models.
\newblock \emph{arXiv preprint arXiv:1701.02386}, 2017.

\end{thebibliography}


\newpage
\appendix
\label{section:app}

\section{Remaining Proofs}

\subsection{Proof of Proposition~\ref{prop:ConcaveGANs}}
\begin{proof}
Look at the first term in the GAN objective, $\E_{\pdata} \log h_\v(\x)$. For a fixed  $x$ we have,
$$\log h_\v(x) = -\log\blr{1+\exp(-\v^\top x)}~,$$
and it can be shown that the above expression is always concave in $\v$
\footnote{For $a\in \reals$, the $1$-dimensional function $Q(a)= -\log\left( 1+\exp(-a)\right)$ is concave. Note that $\log h_\v(x)$ is a composition of $Q$ over a linear function in $\v$, and is therefore concave.}. Since an expectation over concave functions is also concave, this implies the concavity of the first term in $\mathcal H$.

Similarly, look at the second term in the GAN objective, $\E_{\z \in p_\z} \log (1-h_\v(G_\u(\z)))$. 
For a fixed $G_\u(z)$ we have,
$$\log (1-h_\v(G_\u(\z))) = -\log\blr{1+\exp(+v^\top G_\u(\z))}$$
and it can be shown that the above expression is always concave in $\v$. Since an expectation over concave functions is also concave, this implies the concavity of the second term in 
$\mathcal H$.

Thus $\mathcal H$ is a sum of two concave terms and is therefore concave in $\v$.
\end{proof}

\subsection{Proof of Lemma~\ref{lem:NoRegretGamesMinMax}}\label{sec:proof_lemmMinmaxval}
\begin{proof}
Writing explicitly $f_t(\u):= M(\u,\v_t)$ and $g_t(\v):= M(\u_t,\v)$, and plugging these into the regret guarantees of $\A_1,\A_2$, we have,
\begin{align*}
&\sum_{t=1}^T M(\u_t,\v_t)-\min_{\u\in \K_1}\sum_{t=1}^T M(\u,\v_t) \leq B^{\A_1}_T~, \\
&\sum_{t=1}^T -M(\u_t,\v_t)-\min_{\v\in \K_2}\sum_{t=1}^T -M(\u_t,\v) \leq B^{\A_2}_T  ~.
\end{align*}
Summing the above equations and dividing by $T$, we get,
\begin{align} \label{eq:Reg_rel333}
\max_{\v\in\K_2}\frac{1}{T}\sum_{t=1}^T M(\u_t,\v)-\min_{\u\in \K_1}\frac{1}{T}\sum_{t=1}^T M(\u,\v_t) 
 \leq  
 \frac{B^{\A_1}_T}{T} + \frac{B^{\A_2}_T}{T}:=\eps~. 
\end{align}
Next we show that the second term above is always smaller than the minimax value,
\begin{align*} 
\min_{\u\in \K_1}\frac{1}{T}\sum_{t=1}^T M(\u,\v_t) 
& \leq  
\min_{\u\in \K_1}\frac{1}{T}\sum_{t=1}^T \max_{\v\in\K_2}M(\u,\v) \\
& =
\min_{\u\in \K_1} \max_{\v\in\K_2}M(\u,\v) 
\end{align*}
Plugging the above into Equation~\eqref{eq:Reg_rel333}, and recalling $\D_1\sim\text{Uni}\{\u_1,\ldots,\u_t \}$, we get,
\begin{align*} 
\max_{\v\in\K_2}\E_{\u\sim\D_1} M(\u,\v)
 \leq  
 \min_{\u\in \K_1} \max_{\v\in\K_2}M(\u,\v)  +\eps. 
\end{align*}
which concludes the proof.
\end{proof}

\subsection{Proof of the second part of Lemma~\ref{lem:ConcaveFTRL_guarantees} (Stability of FTRL sequence in concave case)}
\label{sec:SecondPart}
\begin{proof}
Here we establish the stability of the FTRL decision rule, $\A_2$, depicted in Theorem~\ref{thm:GuaranteeSemiConcaveGame}.

Note that the following applies to this FTRL objective,
\begin{align}\label{eq:FTRLasProj}
\sum_{\tau=0}^{t-1}\nabla g_\tau(\v_\tau)^\top \v-\frac{\sqrt{T}}{2\eta_0}\|\v\|^2
&=
-\frac{\sqrt{T}}{2\eta_0} \left\| \v - \frac{\eta_0}{\sqrt{T}}\sum_{\tau=0}^{t-1}\nabla g_\tau(\v_\tau)
\right\|^2+C
\end{align}
Where $C$ is a constant independent of $\v$.

Let us denote by $\Pi_{\K_2}$ the projection operator onto $\K_2\subset \reals^n$, meaning,
$$
\Pi_{\K_2}(\v_0) = \min_{\v\in\K_2}\|\v_0-\v\|, \qquad \forall \v_0\in\reals^n
$$
By Equation~\eqref{eq:FTRLasProj} the FTRL rule, $\A_2$, can be written as follows,
\begin{align*}
\v_t 
&=
 \argmin_{\v\in\K_2} 
 \left\| \v - \frac{\eta_0}{\sqrt{T}}\sum_{\tau=0}^{t-1}\nabla g_\tau(\v_\tau)
\right\|^2 \\
&=
\Pi_{\K_2}\left(  - \frac{\eta_0}{\sqrt{T}}\sum_{\tau=0}^{t-1}\nabla g_\tau(\v_\tau)  \right)~.
\end{align*}
The projection operator is a contraction (see e.g,~\cite{hazan2016introduction}), using this together with the above implies,
\begin{align*}
\|\v_{t+1} - \v_t\|
&\leq 
\left\|
\Pi_{\K_2}\left(  - \frac{\eta_0}{\sqrt{T}}\sum_{\tau=0}^{t}\nabla g_\tau(\v_\tau)  \right)
-
\Pi_{\K_2}\left(  - \frac{\eta_0}{\sqrt{T}}\sum_{\tau=0}^{t-1}\nabla g_\tau(\v_\tau)  \right)
\right\|  \\
&\leq
\left\|  - \frac{\eta_0}{\sqrt{T}} \nabla g_t(\v_t)\right\| \\
&\leq
 d_2/\sqrt{2T}~.
\end{align*}
where we used $\|\nabla g_t(\v_t)\|\leq L$ which is due to the Lipschitz continuity of $M$. We also
used $\eta_0 = d_2/\sqrt{2}L$.
\end{proof}

\section{Practical \FTRLGAN Algorithm}
\label{sec:practical_GAN}
The pseudo-code of algorithms $\A_1$ and $\A_2$ is given in Algorithm~\ref{alg:A1_A2}. The algorithm is symmetric for both players and consists as follows. At every step $t$ if we are currently in the switching mode (i.e. $t \mod m == 0$) and the queue is full, we remove a model from the end of the queue, which is the oldest one. Otherwise, we do not remove any model from the queue, but instead just override the head (first) element with the current update.

\begin{algorithm}[h]
\caption{Update queue for Algorithm $\A_1$ and $\A_2$}
\label{alg:A1_A2}
\begin{algorithmic}
\STATE \textbf{Input}: Current step $t$, $m > 0$
\IF {($t \mod m == 0 \text{ and } |\Q|)==K$)}
\STATE $\Q$.remove\_last()
\STATE $\Q$.insert($f_t$)
\STATE $m = m + inc$
\ELSE
\STATE $\Q$.replace\_first($f_t$)
\ENDIF
\end{algorithmic}
\end{algorithm}

We set the initial spacing, $m$, to $\frac{N}{K}$, where $N$ is the number of update steps per epoch, and $K$ is the number of past states we keep. The number of updates per epoch is just the number of the data points divided by the size of the minibatches we use. The default value of $inc$ is 10. Depending on the dataset and number of total update steps, for higher values of $K$, this is the only parameter that needs to be tuned. 
We find that our model is not sensitive to the regularization hyperparameters. For symmetric architectures of the generator and the discriminator (such as DCGAN), for practitioners, we recommend using the same regularization for both players. For our experiments we set the default regularization to 0.1. 

\section{Experiments}
\label{sec:app_experiments}

\subsection{Toy Dataset: Mixture of Gaussians}

The toy dataset consists of a mixture 7 Gaussians with a standard deviation of 0.01 and means equally spaced around a unit circle. \\

The architecture for the generator consists in two fully connected layers (of size 128) and a linear projection to the dimensionality of the data (i.e. 2). The activation functions for the fully connected layers are tanh. The discriminator is symmetric and hence, composed of two fully connected layers (of size 128) followed by a linear layer of size 1. The activation functions for the  fully connected layers are tanh, whereas the final layer uses sigmoid as an activation function. \\

Following ~\cite{metz2016unrolled}, we intialize the weights for both networks to be orthogonal with scaling of 0.8. Adam\cite{kingma2014adam} was used as an optimizer for both the discriminator and the generator, with a learning rate of $1e-4$ and $\beta_1=0.5$. The discriminator and generator respectively minimize and maximize the objective
\begin{equation}
\E_{x \sim \pdata(\x)} -\log(D(x)) - \E_{z \sim \mathcal{N}(0, I_{256})} \log(1-D(G(z))).
\end{equation}

The setup is the same for both models. For \FTRLGAN we use $K=5$ past states with L2 regularization on the network weights using an initial regularization parameter of 0.01. \\

\begin{figure*}[b]
	\begin{center}
          \begin{tabular}{@{}S@{\hspace{2mm}}T@{\hspace{15mm}}T}
          & & \small{\textsc{Target}} \\
          \textsc{GAN} & \includegraphics[width=11.5cm,height=1.5cm,keepaspectratio]{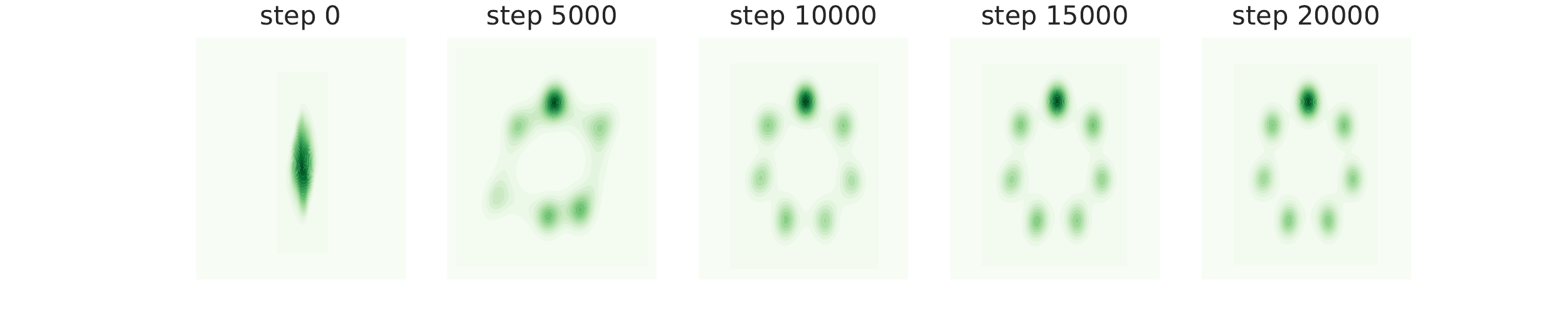}
&
		\includegraphics[width=1.5cm,height=1.3cm,keepaspectratio]{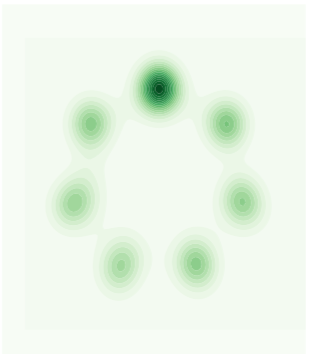} \\
          \textbf{\FTRLGAN} & \includegraphics[width=11.5cm,height=1.5cm,keepaspectratio]{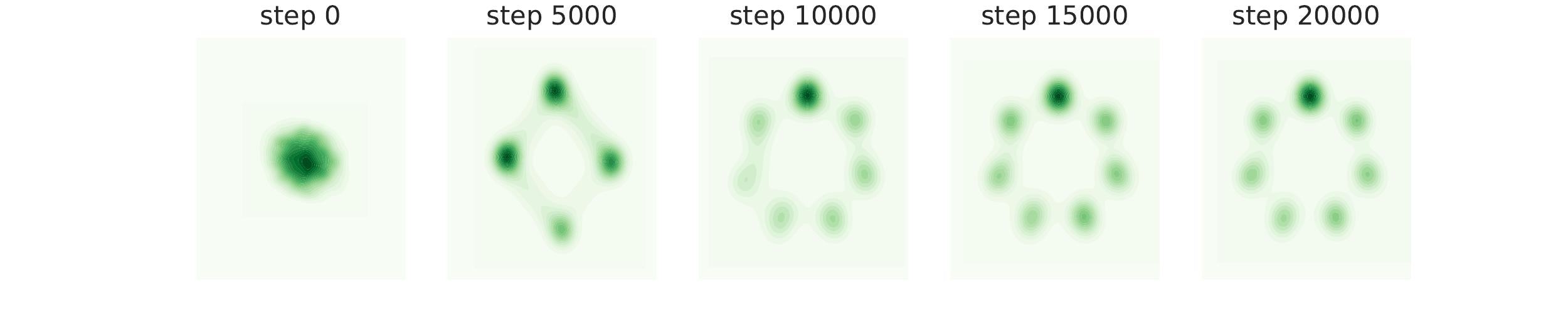}
          
&
		\includegraphics[width=1.5cm,height=1.3cm,keepaspectratio]{Images/Appendix/ToyExperiment/real_data_dim2.png} 
	  \end{tabular}
  \caption{\small{Both GAN and \FTRLGAN converge to the true data distribution when the dimensionality of the noise vector is 2}}
  \label{Appendix:MixtureDim2}
	\end{center}
\end{figure*}

\paragraph{Effect of the latent dimension.} We find that for the case where $z \sim \mathcal{N}(0, I_{256})$, GANs with the traditional updates fail to cover all modes by either rotating around the modes (as shown in \cite{metz2016unrolled}) or converge to only a subset of the modes. However, if we sample the latent code from a lower dimensional space, e.g. $z \sim N(0, I_{2})$, such that it matches the data dimensionality, the generator needs to learn a simpler mapping. We then observe that both GAN and \FTRLGAN are able to recover the true data distribution in this case (see Figure \ref{Appendix:MixtureDim2}).

\begin{figure*}[b]
	\begin{center}
          \begin{tabular}{@{}S@{\hspace{2mm}}T@{\hspace{15mm}}T}
          & & \small{\textsc{Target}} \\
          \textsc{GAN} & \includegraphics[width=11.5cm,height=1.5cm,keepaspectratio]{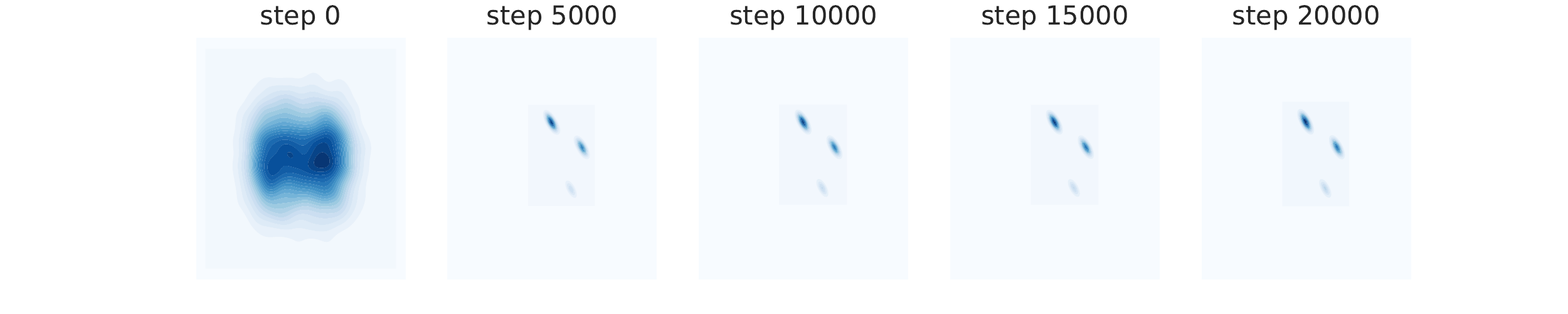}
&
		\includegraphics[width=1.5cm,height=1.1cm,keepaspectratio]{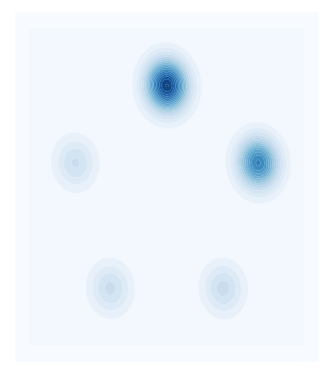} \\
          \textbf{\FTRLGAN} & \includegraphics[width=11.5cm,height=1.5cm,keepaspectratio]{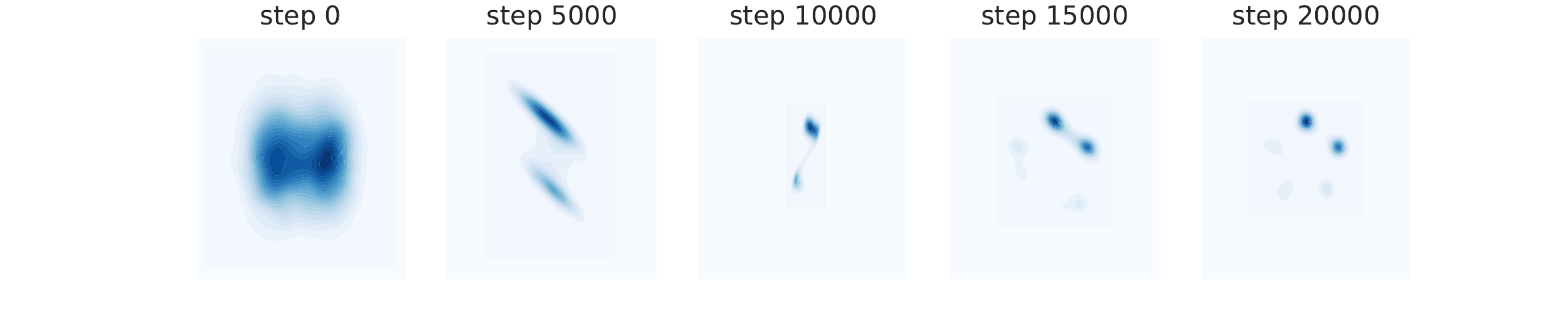}
          
&
	\includegraphics[width=1.5cm,height=1.1cm,keepaspectratio]{Images/Appendix/ToyExperiment/real_data_diff_prob.png} 
	  \end{tabular}
  \caption{\small{\FTRLGAN manages to converge to the true data distribution when two modes have higher sampling probability}}
  \label{Appendix:MixtureDiffModes}
	\end{center}
\end{figure*}

\paragraph{Mode collapse.} We run an additional experiment directly targeted at testing for mode collapse. We sample points $x$ from the data distribution $\pdata$ with different probabilities for each mode. Using the same architectures, we perform an experiment with 5 Gaussian mixtures, again of standard deviation 0.01 arranged in a circle. The probabilities to sample points from each of the modes are [0.35, 0.35, 0.1, 0.1, 0.1]. In this case two modes have higher probability and could potentially attract the gradients towards them and cause mode collapse. \FTRLGAN manages to recover the true data distribution in this case as well, unlike vanilla GANs (Figure \ref{Appendix:MixtureDiffModes}).

\subsection{Augmented MNIST}
We here detail the experiment on the Stacked MNIST dataset. The dataset is created by stacking three randomly selected MNIST images in the color channels, resulting in a 3-channel image that belongs to one out of 1000 possible classes. The architectures of the generator and discriminator are given in Table ~\ref{tbl:GenMNIST} and Table ~\ref{tbl:DiscMNIST}, respectively.

\begin{table}[h!]
\centering
\caption{Stacked MNIST: Generator Architecture}
\label{tbl:GenMNIST}
\begin{tabular}{|c|c|}
\hline
\textbf{Layer}               & \textbf{Number of outputs}           \\ \hline
Input $z \sim N(0, I_{256})$ &                                      \\ \hline
Fully Connected              & 512 (reshape to {[}-1, 4, 4, 64{]} ) \\ \hline
Deconvolution                & 32                                   \\ \hline
Deconvolution                & 16                                   \\ \hline
Deconvolution                & 8                                    \\ \hline
Deconvolution                & 3                                    \\ \hline
\end{tabular}
\end{table}

\begin{table}[h!]
\centering
\caption{Stacked MNIST: Discriminator Architecture}
\label{tbl:DiscMNIST}
\begin{tabular}{|c|c|}
\hline
\textbf{Layer}              & \textbf{Number of outputs} \\ \hline
Convolution                 & 4                          \\ \hline
Convolution                 & 8                          \\ \hline
Convolution                 & 16                         \\ \hline
Flatten and Fully Connected & 1                          \\ \hline
\end{tabular}
\end{table}

We use a simplified version of the DCGAN architecture as suggested by~\cite{metz2016unrolled}. It contains "deconvolutional layers" which are implemented as transposed convolutions. All convolutions and deconvolutions use kernel size of $3 \times 3$ with a stride of 2. The weights are initialized using the Xavier initialization~\cite{glorot2010understanding}. The activation units for the discriminator are leaky ReLUs with a leak of 0.3, whereas the generator uses standard ReLUs.
We train all models for 20 epochs with a batch size of 32, using the RMSProp optimizer with batch normalization. The optimal learning rate for GAN is 0.001, and for \FTRLGAN is 0.01. For all \FTRLGAN models we use regularization of 0.1 for the discriminator and 0.0001 for the generator. The regularization is L2 regularization only on the fully connected layers. For $K=5$, the increase parameter \textit{inc} is set to 50. For K=10, \textit{inc} is 120.

\subsection{CIFAR10 / CelebA}
We use the full DCGAN architecture~\cite{radford2015unsupervised} for the experiments on CIFAR10 and CelebA, detailed in Table ~\ref{tbl:GenDCGAN} and Table \ref{tbl:DiscDCGAN}.

\begin{table}[h!]
\centering
\caption{CIFAR10/CelebA Generator Architecture}
\label{tbl:GenDCGAN}
\begin{tabular}{|c|c|}
\hline
\textbf{Layer}               & \textbf{Number of outputs}           \\ \hline
Input $z \sim N(0, I_{256})$ &                                      \\ \hline
Fully Connected              & 32,768 (reshape to {[}-1, 4, 4, 512{]} ) \\ \hline
Deconvolution                & 256                                   \\ \hline
Deconvolution                & 128                                   \\ \hline
Deconvolution                & 74                                    \\ \hline
Deconvolution                & 3                                    \\ \hline
\end{tabular}
\end{table}

\begin{table}[h!]
\centering
\caption{CIFAR10/CelebA Discriminator Architecture}
\label{tbl:DiscDCGAN}
\begin{tabular}{|c|c|}
\hline
\textbf{Layer}              & \textbf{Number of outputs} \\ \hline
Convolution                 & 64                         \\ \hline
Convolution                 & 128                        \\ \hline
Convolution                 & 256                        \\ \hline
Convolution                 & 512                        \\ \hline
Flatten and Fully Connected & 1                          \\ \hline
\end{tabular}
\end{table}

As for MNIST, we apply batch normalization. The activation functions for the generator are ReLUs, whereas the discriminator uses leaky ReLUs with a leak of 0.3. The learning rate for all the models is 0.0002 for both the generator and the discriminator and the updates are performed using the Adam optimizer. The regularization for \FTRLGAN is 0.1 and the increase parameter \textit{inc} is 10.

\subsubsection{Results on CIFAR10}

\begin{figure}
\centering
\begin{subfigure}
  \centering
  \includegraphics[width=.3\linewidth]{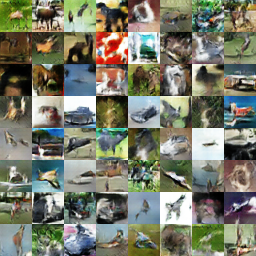}
\end{subfigure}
\quad
\begin{subfigure}
  \centering
  \includegraphics[width=.3\linewidth]{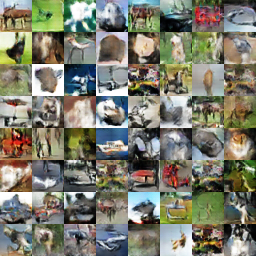}
\end{subfigure}
\caption{Random batch of generated images from GAN after training for 30 epochs (on the left) and \FTRLGAN (K=25) after training for 30 epochs (on the right)}
\label{fig:Cifar2}
\end{figure}

We train for 30 epochs, which we find to be the optimal number of training steps for vanilla GAN in terms of MSE on images from the validation set. Table~\ref{fig:baselines} includes comparison to other baselines. The first set of baselines (given with purple color) consist of GAN where the updates in the inner loop (for the discriminator), the outer loop (for the generator), or both are performed 25 times. The baselines shown with green color are regularized versions of GANs, where we apply the same regularization as in our \FTRLGAN in order to show that the gain is not due to the regularization only. Figure \ref{fig:Cifar2} presents two randomly sampled batches from the generator trained with GAN and \FTRLGAN.

\begin{table}[h!]
\centering
\includegraphics[width=15.5cm,height=12.5cm,keepaspectratio]{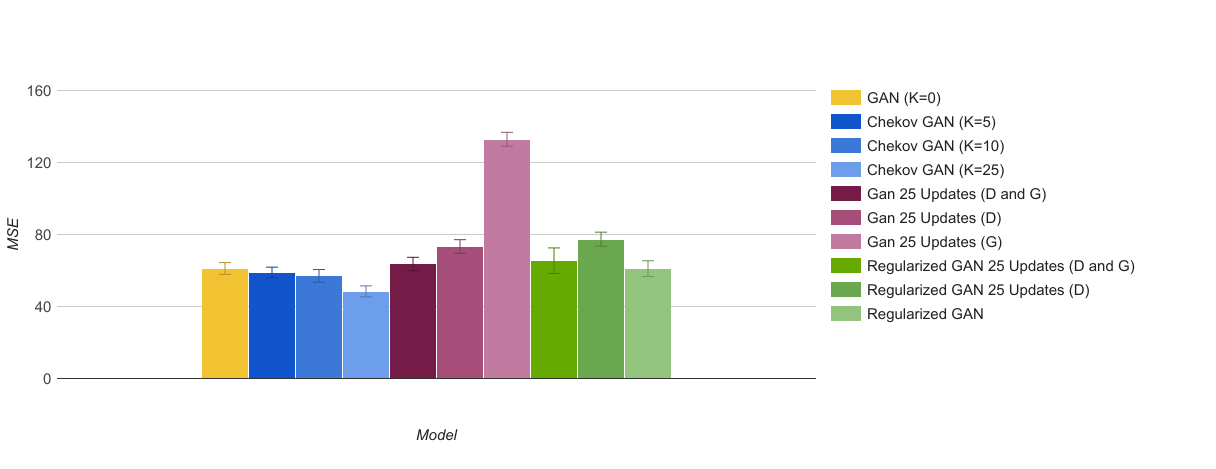}
\caption{CIFAR10: MSE for other baselines on target images that come from the training set. GAN 25 updates indicates that either the generator, the discriminator or both have been updated 25 times at each update step. Regularized GAN is vanilla GAN where the fully connected layers have regularization of 0.05.}
\label{fig:baselines}
\end{table}

\subsubsection{CelebA}

\begin{figure}[ht]
\centering
\begin{subfigure}
  \centering
  \includegraphics[width=.4\linewidth]{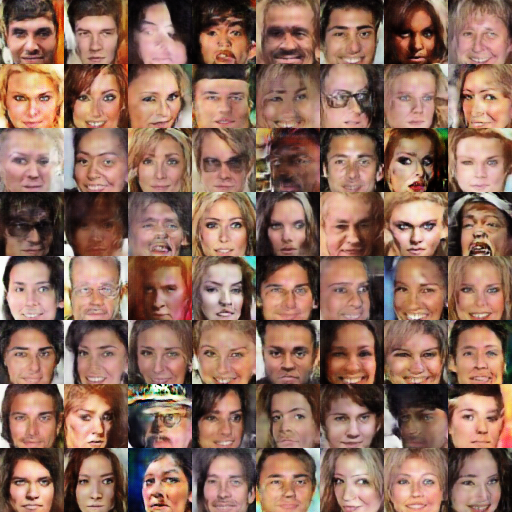}
\end{subfigure}
\quad
\begin{subfigure}
  \centering
  \includegraphics[width=.4\linewidth]{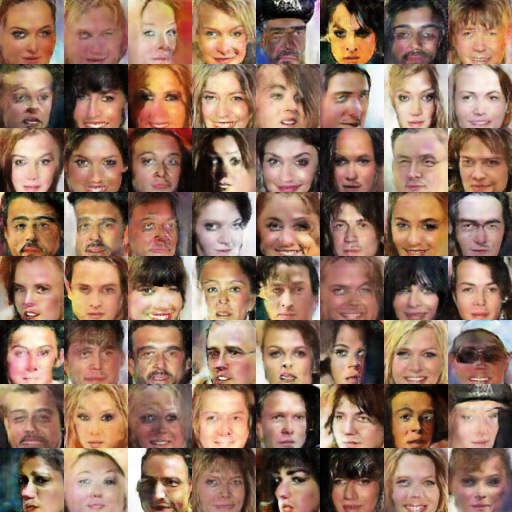}
\end{subfigure}
\caption{Random batch of generated images from GAN (left) and \FTRLGAN ($K=5$) (right) after training for 10 epochs.}
\label{fig:Celeba1}
\end{figure}

All models are trained for 10 epochs. Randomly generated batches of images are shown in Figure~\ref{fig:Celeba1}.

\FloatBarrier 
\FloatBarrier 


\subsubsection{Details about Inference via Optimization on CIFAR10}
\label{sec:inf_optimization}

This approach consists in finding a noise vector $z_{closest}$ that when used as input to the generator would produce an image that is the closest to a target image in terms of mean squared error (MSE):
\noindent\begin{minipage}{.5\linewidth}
\begin{equation*}
  z_{closest} = \argmin_z \text{MSE}(G(z), x_{target})
\end{equation*}
\end{minipage}%
\begin{minipage}{.5\linewidth}
\begin{equation*}
  x_{closest} = G(z_{closest}).
\end{equation*}
\end{minipage}

We report the MSE in image space between $x_{closest}$ and $x_{target}$. This measures the ability of the generator to generate samples that look like real images. A model engaging in mode collapse would fail to generate (approximate) images from the real data. Conversely, if a model covers the true data distribution it should be able to generate any specific image from it. 

\end{document}